\pdfoutput=1
\documentclass[11pt]{article}

\usepackage{ACL2025}

\usepackage{times}
\usepackage{comment}
\usepackage{latexsym}
\usepackage{amsthm}
\usepackage{amsmath,amsfonts,bm,bbm,mathtools,mleftright,nicefrac}
\usepackage{booktabs}
\usepackage{subcaption}
\usepackage{makecell}

\usepackage[capitalize,noabbrev]{cleveref}
\crefname{section}{\S}{\S\S}
\crefname{table}{Tab.}{Tab.}
\crefname{figure}{Fig.}{Figs.}
\crefname{algorithm}{Alg.}{}
\crefname{equation}{Eq.}{Eq.}
\crefname{appendix}{App.}{}
\crefname{theorem}{Theorem}{}
\crefname{restatableTheorem}{Theorem}{}
\crefname{prop}{Proposition}{}
\crefname{definition}{Def.}{}
\crefname{cor}{Corollary}{}
\crefname{observation}{Observation}{}
\crefname{assumption}{Assumption}{}
\crefname{hyp}{Hyp.}{Hypotheses}
\crefformat{section}{\S#2#1#3}
\crefname{namedtheorem}{Hyp.}{Hypotheses}

\usepackage{harmonic}

\usepackage[T1]{fontenc}

\usepackage[utf8]{inputenc}

\usepackage{microtype}

\usepackage{amsfonts}

\usepackage{inconsolata}

\usepackage{graphicx}
\usepackage{xspace}
\usepackage{multirow}

\allowdisplaybreaks

\title{The Harmonic Structure of Information Contours}

\author{
  \textbf{Eleftheria Tsipidi\textsuperscript{\normalfont1}} \quad
  \textbf{Samuel Kiegeland\textsuperscript{\normalfont1}} \quad
  \textbf{Franz Nowak\textsuperscript{\normalfont1}} \quad
  \textbf{Tianyang Xu\textsuperscript{\normalfont2}} \\
  \textbf{Ethan Gotlieb Wilcox\textsuperscript{\normalfont3}} \quad
  \textbf{Alex Warstadt\textsuperscript{\normalfont4}} \quad
  \textbf{Ryan Cotterell\textsuperscript{\normalfont1}} \quad
  \textbf{Mario Giulianelli\textsuperscript{\normalfont1}} \\
  \textsuperscript{1}ETH Zürich \quad
  \textsuperscript{2}TTIC \quad
  \textsuperscript{3}Georgetown \quad
  \textsuperscript{4}UCSD \\
  \texttt{\{\href{mailto:eleftheria.tsipidi@inf.ethz.ch}{eleftheria.tsipidi},
  \href{mailto:samuel.kiegeland@inf.ethz.ch}{samuel.kiegeland},
  \href{mailto:franz.nowak@inf.ethz.ch}{franz.nowak},} \\
  \texttt{\href{mailto:ryan.cotterell@inf.ethz.ch}{ryan.cotterell},
  \href{mailto:mario.giulianelli@inf.ethz.ch}{mario.giulianelli}\}@inf.ethz.ch},\\
  \texttt{\href{mailto:sallyxu@ttic.edu}{sallyxu@ttic.edu},
  \href{mailto:ethan.wilcox@georgetown.edu}{ethan.wilcox@georgetown.edu},
  \href{mailto:awarstadt@ucsd.edu}{awarstadt@ucsd.edu}}\\
}
\begin{document}
\maketitle
\begin{abstract}
The uniform information density (UID) hypothesis proposes that speakers aim to distribute information evenly throughout a text, balancing production effort and listener comprehension difficulty. However, language typically does not maintain a strictly uniform information rate; instead, it fluctuates around a global average. These fluctuations are often explained by factors such as syntactic constraints, stylistic choices, or audience design. In this work, we explore an alternative perspective: that these fluctuations may be influenced by an implicit linguistic pressure towards periodicity, where the information rate oscillates at regular intervals, potentially across multiple frequencies simultaneously.
We apply harmonic regression and introduce a novel extension called time scaling to detect and test for such periodicity in information contours. Analyzing texts in English, Spanish, German, Dutch, Basque, and Brazilian Portuguese, we find consistent evidence of periodic patterns in information rate. Many dominant frequencies align with discourse structure, suggesting these oscillations reflect meaningful linguistic organization. Beyond highlighting the connection between information rate and discourse structure, our approach offers a general framework for uncovering structural pressures at various levels of linguistic granularity.

\vspace{.11em}
\hspace{1.25em}\includegraphics[width=1.25em,height=1.25em]{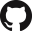}{\hspace{.75em}\parbox{\dimexpr\linewidth-2\fboxsep-2\fboxrule}{\url{https://github.com/rycolab/harmonic-surprisal}}}
\end{abstract}

\section{Introduction}
\begin{figure*}
    \centering
    \includegraphics[width=\linewidth]{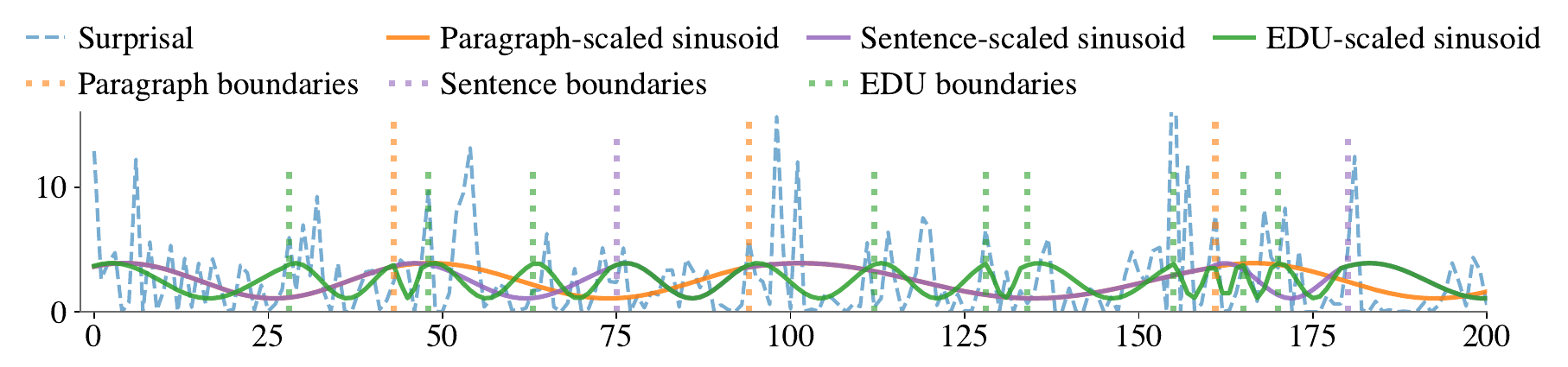}
    \caption{\textbf{Illustration of Harmonic Regression on Surprisal Contours}. Surprisal contours, unit boundaries, and first-order sinusoids for the first $200$ tokens from a Wall Street Journal article (document \texttt{wsj\_1111} in the English RST Discourse Bank). Time scaling (\Cref{sec:time-scaling}) is applied according to the lengths of elementary discourse units (EDUs), sentences, and paragraphs. Here, we set the coefficients of the sinusoids to $1$ for illustrative purposes. See \cref{fig:predictive-harmonics-vis-ma_00059a} and \cref{sec:sin-vis} for realistic decompositions.
    }
    \label{fig:time-scaling}
    \vspace{-15pt}
\end{figure*}

Studying the rate at which speakers transmit information has been a long-standing topic of interest in linguistics and cognitive science \citep[\emph{inter alia}]{shannon1948mathematical, genzel-charniak-2002-entropy, bell2003effects,xu-reitter-2018-information,giulianelli-fernandez-2021-analysing}.
From an information-theoretic perspective, effective communication involves striking a balance between a rate sufficiently low for the receiver to successfully decode the intended message and yet sufficiently high for the sender to reduce their effort \citep{zipf1949human,clark-1986-referring,aylett1999,aylett-turk-2004,gibson2019efficiency}. In this context, information is often quantified as Shannon surprisal, i.e., the negative log probability, of the unit being communicated within its context.
As evident from visualizing surprisal values across a text---see, for example, \Cref{fig:time-scaling} for token-level surprisals estimated with a Transformer language model---information rate fluctuates harmonically throughout the discourse. 

To predict and explain fluctuations in the surprisal of units, prior work has examined their relationship with a unit's position within elements of the discourse structure, such as paragraphs \citep{genzel-charniak-2003-variation}, topic episodes \citep{qian2011topic,xu-reitter-2016-entropy}, and dialogue-specific contextual units \citep{giulianelli-etal-2021-information,maes2022shared}. 
While these studies independently establish links between aspects of discourse context and information rate, a comprehensive framework for investigating when and how a unit's position within its contextual structure affects its information remains an open question. More broadly, no overarching theory yet accounts for harmonic structure in global information contours.

One influential framework for understanding the relationship between information rate and \emph{local} context is the \defn{uniform information density (UID)} hypothesis \citep{fenk1980konstanz,aylett-turk-2004,levy-jaeger-2007-reduction, meister-etal-2021-revisiting}. 
This hypothesis posits that, within the constraints of grammar, speakers tend to distribute information as evenly as possible.
UID accounts for production choices within narrow syntactic and discourse contexts;
in particular, in environments that allow for multiple alternative formulations, speakers favor those that achieve greater information uniformity \citep{jaeger2010redundancy,maholwald-2013-info,torabi-asr-demberg-2015-uniform}.
However, when UID is stretched to larger contextual units, its explanatory power weakens.\looseness=-1

At a \emph{global} level, the UID hypothesis has been taken to imply that each linguistic unit contributes a constant amount of information throughout a discourse, which corresponds to a fully rational use of the communication channel~\citep{genzel-charniak-2002-entropy}. 
Empirical findings, however, show that surprisal curves over discourse are rarely static \citep{xu-reitter-2016-entropy,giulianelli-fernandez-2021-analysing,verma-etal-2023-revisiting}; instead, surprisal fluctuates within a bounded range; see also \Cref{fig:time-scaling}. 
This suggests the pressure for uniformity is counterbalanced by competing functional constraints on communication beyond grammaticality, which become increasingly influential in longer stretches of text and dialogue. 
Modulating surprisal within discourse can indeed serve various functions, such as adhering to aesthetic or stylistic conventions \citep{lewis1894history}, maintaining listener engagement and supporting comprehension \citep{cervantes1992effects} reducing the cognitive demands of real-time production \citep{bergey2024producing}, and enhancing task success in cooperative interactions \citep{yee-etal-2024-efficiency}. 
The way in which these global pressures shape the harmonic structure of information contours is not yet well understood.\looseness-1

To fill this gap,
this paper introduces the \defn{harmonic surprisal (HS)} hypothesis. 
We hypothesize that surprisal contours can be globally described as a mixture of periodic trends, and that the lengths of the different periods align with structural units of varying granularity.
The HS hypothesis is a refinement of the recently introduced \defn{structured context (SC)} hypothesis~\citep{tsipidi-etal-2024-surprise}.
Both hypotheses posit that a unit's surprisal can be predicted from the unit's position within the contextual structure.
However, the HS hypothesis establishes a specific relationship between position and surprisal rate---one that is governed by a periodic function.
To operationalize the HS hypothesis, we propose a simple modification of harmonic regression. 
Harmonic regression is a variant of linear regression that models dependent variables as a combination of sinusoidal components; it offers a convenient way of examining whether surprisal contours exhibit periodic trends. 
Our modification allows us to embed hypotheses about relevant contextual structures directly into the statistical modeling procedure. Specifically, scaling the sinusoidal predictors by the length of a given structural unit, we are able to test whether the periodic trends of surprisal contours align with the boundaries of that unit. 
In doing so, this approach enables us to uncover interpretable harmonic structures that reflect the underlying contextual organization.

Building on \citeposs{tsipidi-etal-2024-surprise} findings, we focus our analysis on sentences, paragraphs, and elementary units of the rhetorical discourse structure \citep{mann-thompson-rst-1988}.
However, the flexibility of the proposed harmonic regression approach also allows us to observe the role of smaller units, whose influence on surprisal fluctuations is modeled by sinusoidal components with higher frequencies.
Our analyses on English, Spanish, German, Dutch, Basque, and Brazilian Portuguese texts provide consistent evidence for periodicity in surprisal contours. 
This evidence for periodicity is particularly pronounced when we time-scale predictors to align with the boundaries of elementary discourse units (EDUs), with first-order sinusoids---those exactly corresponding to the EDU spans---having the highest amplitude.
Overall, our findings indicate that discourse structure influences surprisal dynamics in text, with periodic patterns in surprisal contours emerging in alignment with discourse structure constituents.\looseness-1

\section{Information Contours}
The standard information-theoretic approach to analyzing the rate at which speakers transmit information in text and speech is to track \defn{information contours}---time series representing per-unit information throughout the linguistic signal \cite{genzel-charniak-2002-entropy,keller-2004-entropy,xu-reitter-2016-entropy,giulianelli-fernandez-2021-analysing}.
While alternative measures of information may be suitable to such analyses \cite[][\textit{inter alia}]{rabovsky2018modelling,aurnhammer2019evaluating,giulianelli-etal-2023-information,giulianelli-etal-2024-generalized,giulianelli-etal-2024-incremental,meister-etal-2024-towards,li2024information}, this work adheres to the classical information-theoretic model, using surprisal \cite{shannon1948mathematical} as a measure of information.
The following sections introduce key notation and concepts, along with several prominent hypotheses regarding the functional pressures shaping surprisal contours.

\subsection{Alphabets, Strings, and Documents}
An \defn{alphabet} $\alphabet$ is a non-empty set of symbols.
A \defn{string} $\str$ over alphabet $\alphabet$ is a finite sequence of symbols $\str = \sym_1 \cdots \sym_N$, 
where $\sym_1, \ldots, \sym_N \in \alphabet$. 
The string's length $N$ is denoted as $\card{\str}$ and the empty string as $\varepsilon$. 
The set of all strings composed of symbols in $\alphabet$ is denoted as $\Strings$. 
Given a string $\str$ of length $\card{\str} \geq t$, $\str_{<t}$ is the string composed of the first $t\!-\!1$ symbols of $\str$.
We write $\str \preceq \str'$ if $\str$ is a prefix of $\str'$ and denote the concatenation of two strings $\str$, $\str'$ as $\str \str'$.
We define a \defn{document}, such as a full text or dialog, as a string $\str \in \Strings$.\footnote{
	To encode a text or a dialog as a string, sentence breaks, paragraph breaks, turn transitions, and other markers of conventional document structure must be expressible in $\Strings$.
}

\subsection{Language Models}
Given an alphabet $\alphabet$, a \defn{language model} $\plm$ is a probability distribution over strings $\Strings$ composed of symbols from the alphabet.
The \defn{prefix probability} under language model $\plm$ is defined as
\begin{align}
	\label{eq:prefix-prob}
	\prefixlm(\str) \defeq \sum_{ \str' \in \Strings } \indicator{\str \preceq \str'}  \plm(\str').    
\end{align}
\Cref{eq:prefix-prob} is the probability that a string has $\str$ as prefix.
The prefix probability can be used to define the conditional prefix probability of a target string $\str'$ given its preceding context $\str$:
\begin{subequations}
	\begin{align}
		\prefixlm(\str' \mid \str) 
		&\defeq 
		\frac{\prefixlm(\str \str')}{\prefixlm(\str)}, \\
		\prefixlm(\eos \mid \str) 
		&\defeq \frac{\plm(\str)}{\prefixlm(\str)}.
		\label{eq:conditional-prefix-prob}
	\end{align}
\end{subequations}
\Cref{eq:conditional-prefix-prob} is the conditional prefix probability of the \underline{e}nd-\underline{o}f-\underline{s}tring event $\eos$ given a context, i.e., the probability that, if the language model $\plm$ is to generate the string $\str$, then it will only generate $\str$ and not continue it in the manner of $\str \str'$.\looseness-1

Every language model can be expressed in autoregressive form by decomposing the probability of a string as the product of conditional probabilities of each of its symbols, followed by 
$\eos$:
\begin{equation}
	\label{eq:autoregressive}
	\plm(\str) = \prefixlm(\eos \mid \str) \prod_{t = 1}^{|\str|} \prefixlm\mleft(\sym_t \mid \str_{<t} \mright),
\end{equation}
where each conditional distribution $\prefixlm\mleft(\cdot \mid \str{<t}\mright)$ is a probability distribution over $\alphabet \cup \{\eos\}$.

\paragraph{Modeling the Human Language Model.} Many modern language models are defined via the product in \Cref{eq:autoregressive}, with each conditional prefix probability derived from a parametric model, such as a neural network. 
In this paper, we use neural network models as a proxy for a particular hypothetical construct model, i.e., the human language model, which is unknown.

\subsection{Surprisal}
\label{sec:surprisal_notation}
Given a document $\str = \sym_1 \cdots \sym_T$, the surprisal of a unit $\sym_t$ given its preceding context $\str_{<t}$ is defined as the negative logarithm of the unit's conditional probability:
\begin{equation}
\label{eq:surprisal}
\surprisal(\sym_t; \str_{<t}) \defeq - \log \prefixlm(\sym_t \mid \str_{<t}),
\end{equation}
where $\prefixlm$ is the prefix probability of a language model $\plm$.
Note that the surprisal at the beginning of the string is given by $\surprisal(w_1)\defeq -\log\prefixlm(w_1 \mid \varepsilon)$ and at the end of the string by $\surprisal(\eos; \str) \defeq - \log \prefixlm(\eos \mid \str)$.
Here, for simplicity, we assume that the set of units of interest corresponds to the alphabet of the language model; however, this need not be the case \cite{giulianelli-etal-2024-proper,vieira-2024-token-to-char-lm}.
The \defn{surprisal contour} of a document $\str = \sym_1 \cdots \sym_T$ is defined as
\begin{align}
    \strsurprisals=[\surprisal(\sym_1),\surprisal(\sym_2;\sym_1), \ldots, \surprisal(\eos; \str)]^\top.
\end{align}

\paragraph{The Role of Surprisal in Psycholinguistics.}
Beyond measuring the information content of linguistic units, surprisal plays an important role in psycholinguistic theory as a measure of processing effort in human language comprehension.
In particular, surprisal theory posits that the effort incurred by a comprehender in processing a unit is in a logarithmic relationship with its contextual probability or, equivalently, that it is proportional to the unit's surprisal.
This relationship has been confirmed empirically by a large body of work using neural and behavioral measurements of processing effort in reading and listening \citep[][\textit{inter alia}]{fernandez-monsalve-etal-2012-lexical,smith-levy-2013,frank2015erp,goodkind2018predictive,shain2020fmri,shainetal24,schrimpf2021,wilcox-etal:2020-on-the-predictive-power,wilcox-2023-testing,wallbridge2022investigating,xu-etal-2023-linearity,Huber2024-vt}.
Speakers' modulation of surprisal has also been shown to explain a variety of phenomena in language production \cite{bell2003effects,aylett-turk-2004,aylett2006language,levy-jaeger-2007-reduction,frank2008speaking,jaeger2010redundancy,futrell-2023-production,yee-etal-2024-efficiency}.
In addition to examining the relationship between individual surprisal values $\surprisal(\sym_t; \str_{<t})$ and linguistic constructs or phenomena, another insightful approach in psycholinguistics is to analyze the broader dynamics of surprisal contours $\strsurprisals$.
The following sections discuss key hypotheses about the functional pressures that influence the shape of global surprisal contours.\looseness-1

\subsection{Uniform Information Density}
\label{sec:contours-uid}
One of the most prominent hypotheses regarding the shape of surprisal contours is the uniform information density (UID) hypothesis \citep{fenk1980konstanz,aylett-turk-2004,levy-jaeger-2007-reduction}. 
UID has been proposed as a constraint across multiple levels of the linguistic hierarchy, affecting consonant deletion \citep{cohenpriva2015informativity}, syllable duration \citep{aylett-turk-2004}, word abbreviation \citep{maholwald-2013-info}, syntactic reduction \citep{levy-jaeger-2007-reduction}, and discourse as a whole \citep{genzel-charniak-2002-entropy}.
When extended to global contexts, the UID hypothesis can be expressed as follows.
\begin{hypothesis}[Uniform Information Density; UID]\label{hypothesis-uid}
    Subject to the constraints of the grammar, speakers optimize their linguistic signals such that the surprisals $\strsurprisals$ are distributed as uniformly as possible throughout a document $\str$.
\end{hypothesis} 
There are several ways to operationalize the uniformity of the information contour $\strsurprisals$. Uniformity can be expressed either through local variance, where the surprisal of adjacent units is evenly distributed, or through global variance, where surprisal tends to regress toward a global mean \cite{collins2014information}.
Comparing these two operationalizations, \citet{meister-etal-2021-revisiting} and \citet{giulianelli-fernandez-2021-analysing} find stronger evidence for information uniformity on a global scale---whether considering words or entire utterances as linguistic units---supporting the notion that, at the discourse level, UID is better understood as a regression toward a mean information rate. Nonetheless, the pressure for surprisal to regress to a mean does not fully account for the variability and internal structure of surprisal contours.
Specifically, one might expect fluctuations in surprisal contours not merely to represent noise around the global mean but instead to reflect linguistic structures of varying granularity, ranging from collocations and syntactic constructions to broader discourse organization.

\subsection{The Structured Context Hypothesis}
\label{sec:contours-sch}
Regular fluctuations in surprisal values have been observed in empirical studies at nearly every level for which UID has been claimed, including the character level \citep{elman1990finding}, the syntactic level \citep{slaats2024lexical},
and the discourse level \citep{xu-reitter-2016-entropy,xu-reitter-2018-information,giulianelli-fernandez-2021-analysing,maes2022shared,verma-etal-2023-revisiting}.
Taken together, these studies indicate that there are pressures beyond UID influencing the shape of surprisal contours and that the deviations of surprisal away from a global base rate follow a pattern that is predictable from the unit's position within its containing structural units.
\citet{tsipidi-etal-2024-surprise} propose the structured context (SC) hypothesis as a refinement of the UID hypothesis.
The SC hypothesis explicitly describes the relationship between surprisal contours and the hierarchical structure of a document.\looseness=-1
\begin{hypothesis}[Structured Context; SC]
\label{hyp:hsc}
  The components $\surprisal(\sym_t; \str_{<t})$ in the surprisal contour $\strsurprisals$ of a document $\str$ are (partially) determined by the position of $\sym_t$ within the hierarchy of $\str$'s constituent structural units.\looseness=-1
\end{hypothesis}
In other words, this hypothesis posits that we should be able to predict surprisal contours from features that describe the hierarchical structure of language at different levels.
\citet{tsipidi-etal-2024-surprise} test this at the discourse level using features such as the relative position of $\sym_t$ within a higher-level structure (e.g., a sentence or a rhetorical discourse unit) and the position of structural units that contain $\sym_t$ within their parent node as predictors in a linear model. 
They find that both shallower and deeper types of hierarchical features are significant predictors of surprisal contours.
However, the predictive power of their models is moderate, leaving open the possibility that refinements to the structural features and our assumptions about their relationships to surprisal could better explain fluctuations in information contours. 

\subsection{The Harmonic Surprisal Hypothesis}\label{sec:bottom-up-harmonic-surprisal}
To offer a more precise account of global surprisal fluctuations, we propose a refinement of the UID and SC hypotheses by restricting the set of possible explanations to those that inherently capture the oscillatory nature of surprisal contours.
\begin{hypothesis}[Harmonic Surprisal; HS]
\label{hyp:hsh}
  The components $\surprisal(\sym_t; \str_{<t})$ in the surprisal contour $\strsurprisals$ of a document $\str$ vary periodically, with periods that correspond to the boundaries of structural units within $\strsurprisals$. 
\end{hypothesis}
This can be understood as stating that the predictability of surprisal contours posited by the SC hypothesis arises from a certain degree of regularity in surprisal fluctuations and that such regularity can be better described as a mixture of periodic patterns.
Our experiments examine whether the periods contributing to this mixture correspond to the span of discourse units.
Furthermore, the reason we refer to \Cref{hyp:hsh} as an elaboration of UID, rather than a competing hypothesis, is that the existence of periodic structure in $\strsurprisals$ does not contradict that surprisal values should be evenly distributed locally, e.g., with adjacent surprisals $\surprisal(\sym_t;\str_{<t})$ and $\surprisal(\sym_{t+1};\str_{<t+1})$ of similar magnitude, nor does it contradict a global notion of uniformity where surprisal values tend to accumulate around the mean surprisal in $\strsurprisals$.\looseness=-1

\section{Harmonic Regression Modeling}
\Cref{hyp:hsh} motivates the search for a statistical method that can automatically discover periodic structure in the information contour $\strsurprisals$.
In this work, we adopt a time series modeling perspective, specifically using \defn{harmonic regression}, a parameterization of linear regression that incorporates sinusoids as independent variables. 
We define the global per-unit surprisal as our dependent variable and BPE-segmented tokens as our base units $\sym_t$.
Harmonic regression is based on the principle that any periodic function can be approximated using a sum of sine and cosine functions.
Beyond detecting periodicity, harmonic regression enables us to explore whether any fluctuations in surprisal we find align with the structural units believed to influence them. For other approaches to surprisal contour modeling, see \Cref{sec:appendix_other_surprisal_modeling}.\looseness=-1

\subsection{Harmonic Regression}
\label{sec:harmonic_regression}
Harmonic regression models a periodic function $\periodicf(t)$ as a linear combination of sine and cosine components at integer multiples of its fundamental frequency:
\begin{equation}
\begin{aligned}
\label{eq:harmonic_regression}
    \periodicf(t) = \beta_0 + \sum_{\component=1}^{\order} & ~ \left(\beta_{1,\component}\cdot \sin \left(\frac{\component2\pi t}{\tslength}\right)\right. \\ &+ \left. \beta_{2,\component} \cdot \cos \left(\frac{\component2\pi t}{\tslength}\right)\right),    
\end{aligned}
\end{equation}
where $\order$ is the order of the model, i.e., the number of harmonic components, $\tslength$ is the length of the longest period, and $\frac{\component}{\tslength}$ is the frequency. 
The coefficient $\beta_0$ controls the vertical shift of the series while $\beta_{1,\component}$ and $\beta_{2,\component}$ scale the contribution of the sine and cosine of the harmonic component $\component$, yielding the model parameter vector $\vbeta \in \R^{2\order+1}$,
\begin{align}
    \vbeta = [\beta_0, \beta_{1,1}, \dots, \beta_{1,K}, \beta_{2,1}, \dots,\beta_{2,\order}]^{\top}.
\end{align}
The amplitude of the $\component^{\text{th}}$ harmonic component is given by $\amp_\component = \sqrt{\beta_{1,\component}^2 + \beta_{2,\component}^2}$, capturing the strength of that frequency component.
The parameters are then estimated by minimizing the ordinary least-squares objective.
Harmonic regression identifies the combination of sinusoids that best predicts the shape of a surprisal curve in a fully unsupervised manner. 
However, it does not offer a way to examine the influence of linguistic structures (e.g., paragraphs or sentences) that we might a priori expect to be predictive of periodic surprisal patterns.
To test hypotheses about such structures, we introduce an additional scaling mechanism in the time domain of the signal.

\subsection{Time Scaling}
\label{sec:time-scaling}
Time scaling adjusts the period of the sinusoids in the harmonic regression to account for the span of different structural elements containing the base unit $\sym_t$.
This results in a modified summation term for the $\component^{\text{th}}$ harmonic component (cf.~\Cref{eq:harmonic_regression}):
\begin{equation}
    \beta_{1,\component}\cdot \sin \left(\frac{\component2\pi t}{\unitlen_t}\right) +  \beta_{2,\component} \cdot \cos \left(\frac{\component2\pi t}{\unitlen_t}\right),
\end{equation}
where $\unitlen_t$ is the length of the structural unit containing $\sym_t$.
When considering the entire document $\str$, i.e., the coarsest unit containing $\sym_t$, and normalizing by its length $|\str|$, this corresponds to adjusting periods to the relative rather than absolute position of $\sym_t$. This serves as our reference condition.
Time scaling can be applied to linguistic structures of varying granularity, from syntactic constructions and multi-word expressions to larger discourse structures.
In the present work, we focus on discourse structural predictors---in particular, paragraphs, sentences, and elementary discourse units (EDUs, i.e., the smallest meaningful units of discourse in rhetorical structure theory; \citealp{mann-thompson-rst-1988}). See \Cref{sec:datasets} for more details on these predictors.
For every $\sym_t$, we scale the periods of the sinusoids by the length $\unitlen_t$ (measured as the number of tokens) of $\sym_t$'s containing paragraph, sentence, and EDU.
Furthermore, for each granularity level, we set $\order$ to the length of the longest unit in the training set of each cross-validation fold. This ensures that our harmonic components represent periods ranging from one base unit $\sym_t$ to the full length of the longest unit. 
For an example of first-order sinusoids scaled to our three discourse structures of interest, see \Cref{fig:time-scaling}.
Time scaling allows us to test our HS hypothesis by observing whether structure-specific sinusoids are significant predictors of surprisal contours. 

\subsection{Feature Selection and Significance}
\label{ssec:regularization-baseline}
We fit linear models including baseline features and the aforementioned time-scaled sinusoids using 10-fold cross-validation.\footnote{
    Baseline features include the number of characters in~$\sym_t$, previous timestep surprisal $\surprisal(\sym_{t-1}; \str_{<t-1})$, relative position of $\sym_t$ in $\str$, and boolean feature vectors indicating whether $\sym_t$ is within windows of 1, 2, and 4 tokens distance from a structural boundary. The latter are included to test whether harmonic features capture periodicity beyond what can be explained by changes in surprisal at structural boundaries.\looseness-1
}
We perform feature selection using $L_1$ regularization (see \Cref{sec:reg-details}), and we then refit the model using only the features with non-zero coefficients in the regularized fit. We use one-way ANOVA to assess statistical significance for the remaining features by comparing each harmonic order (sine and cosine) against a baseline model that includes only the non-sinusoid baseline features. For more details, see \Cref{sec:baseline-details}.

\section{Data}
We test for periodic structure in surprisal contours across six languages:
English, Spanish, German, Dutch, Basque, and Brazilian Portuguese.
Surprisal contours are obtained using Transformer-based language models as estimators.
\begin{table*}[ht!]
\small
\centering
\setlength{\tabcolsep}{4pt}
\begin{tabular}{lcccccc}
\toprule
 & \textbf{English} & \textbf{Spanish} & \textbf{German} & \textbf{Dutch} & \textbf{Basque} & \textbf{Brazilian Portuguese} \\
\midrule
Baseline  & $9.91\pm.43$ & $14.63\pm.47$ & $12.43\pm.25$ & $9.32\pm.79$ & $9.00\pm.55$ & $9.62\pm.81$ \\
\midrule
Document-scaled & $9.92\pm.44$ & $\mathbf{13.52\pm.38}$* & $12.29\pm.26$* & $9.60\pm.81$ & $9.17\pm.52$ & $9.80\pm.78$ \\
EDU-scaled & $\mathbf{9.46\pm.40}$* & $13.83\pm.45$* & $\mathbf{11.31\pm.29}$* & $\mathbf{8.73\pm.74}$* & $\mathbf{8.67\pm.55}$* & $\mathbf{9.07\pm.83}$* \\
Sentence-scaled & $9.55\pm.42$* & $14.17\pm.45$* & $11.56\pm.30$* & $8.92\pm.72$* & $8.85\pm.56$* & $9.32\pm.84$* \\
Paragraph-scaled & $9.73\pm.43$* & $14.40\pm.41$* & $12.23\pm.27$* & $9.31\pm.75$ & $9.24\pm.53$ & $9.55\pm.85$ \\
\midrule
All & $9.37\pm.40$* & $13.09\pm.35$* & $11.37\pm.33$* & $9.22\pm.74$ & $9.08\pm.53$ & $9.42\pm.80$* \\
\bottomrule
\end{tabular}
\caption{Mean and standard deviation of validation MSE across ten cross-validation folds for each harmonic regression model and language. Bolded values indicate the lowest MSEs, excluding the maximal (All) model. Asterisks (*) denote models that significantly outperform the baseline according to a one-sided paired $t$-test ($\significance < 0.001$); see \Cref{sec:significance-baseline} for further details on significance testing.
}
    \label{tab:val_mse}
    \vspace{-5pt}
\end{table*}
\begin{figure*}[ht!]
\centering
    \includegraphics[width=\textwidth]{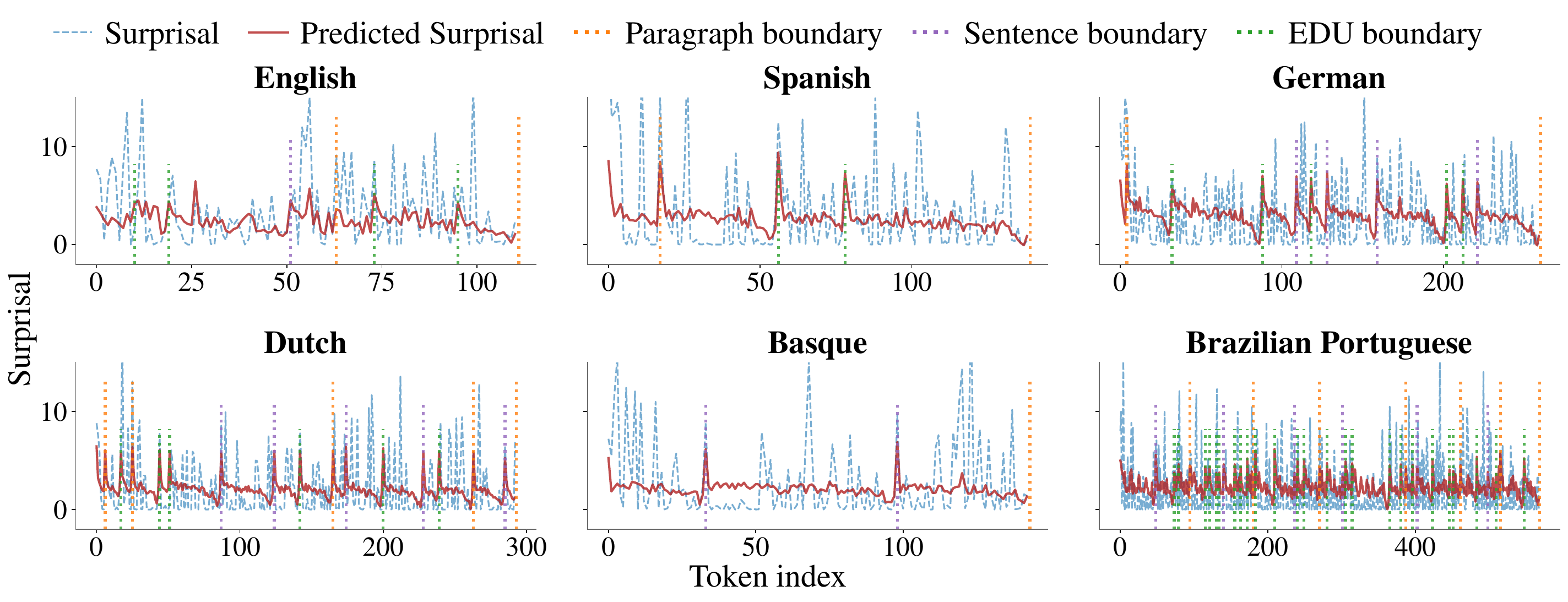}
    \caption{\textbf{Predicted vs. Observed Surprisal Curves for EDU-scaled Sinusoids}. Each panel shows predictions for one document: English (\texttt{wsj\_0605}), Spanish (\texttt{as00007}), German (maz-1818), Dutch (\texttt{AD14\_CarpeDiem}), Basque (GMB0002-GS), Brazilian Portuguese (\texttt{D2\_C38\_Estadao}). For other scaling methods, see \Cref{sec:prediction-vis}.\looseness-1
    }
    \label{fig:predicted-curves-edu}
    \vspace{-5pt}
\end{figure*}

\subsection{Datasets}
\label{sec:datasets}
Following \citet{tsipidi-etal-2024-surprise}, we base our analysis on datasets annotated according to Rhetorical Structure Theory (RST), a widely recognized formalism for analyzing discourse structure which originated from early work in text generation and later developed into a linguistic theory. 
We use six RST-annotated corpora covering English, Spanish, German, Dutch, Basque, and Brazilian Portuguese \citep{carlson-etal-2001-building, carlson-marcu-2001-discourse, da-cunha-etal-2011-development, stede-2004-potsdam, stede-neumann-2014-potsdam, vliet-etal-2011-building, redeker-etal-2012-multi, iruskieta-etal-2013-RSTBasque, Cardoso-etal-2011}.
RST segments texts into recursively nested spans linked by rhetorical relations, and its basic units of analysis are elementary discourse units (EDUs)---the smallest communicative segments in a discourse tree, which convey complete propositions and serve as the fundamental building blocks of larger logical and rhetorical structures.\footnote{
    EDUs often align with clauses, but there are also instances where an EDU may contain more than a single clause, such as clauses where the subject or object of the main verb is also a clause \cite{carlson-marcu-2001-discourse}.
}

In addition to RST-based discourse segmentation, we also consider sentences and paragraphs as conventional prose structures.
Each dataset is thus processed by segmenting documents into paragraphs, sentences, and EDUs. \Cref{tab:datstats} summarizes the total counts of these units, while \Cref{tab:stats} provides an overview of the mean and median number of EDUs per sentence, paragraph, and document; both tables and additional information on the datasets can be found in \Cref{sec:dataset-stats}.\looseness-1

\subsection{Surprisal Estimation}
We compute global per-token surprisal according to \Cref{eq:surprisal} for every document in the respective dataset. For each language, we select a dedicated open-weight LLM, fine-tuned on data in that language.\footnote{For an overview of the models, see \Cref{tab:model-stats}.}

\section{The Harmonics in Surprisal Contours}
\label{sec:empirical-findings}
We start by evaluating the overall predictive power of harmonic regression for surprisal contours. We fit five 
models: one for each type of structure---i.e., sinusoids scaled to the document, EDU, sentence, or paragraph level---and a maximal model that includes sinusoids scaled by all structures simultaneously. All models include the baseline predictors, as described in \Cref{ssec:regularization-baseline}. Predictive power is reported as the mean validation MSE over 10-fold cross-validation, with results summarized in \Cref{tab:val_mse}. 

We find that scaling periods by EDU length leads to the lowest MSE across all languages except for Spanish. For the other five languages, EDU-scaling yields similar or better MSE than the maximal model (All).\footnote{
In \Cref{sec:permuted_surprisal}, we repeat the experiments with randomly permuted surprisal values, which results in higher MSE and no notable differences between scaling methods. 
} 
We test for statistical significance against a model that includes only baseline predictors, and find that all models incorporating EDU- or sentence-scaled sinusoids significantly outperform this baseline.
To further provide a visual assessment of the best model's fit, in \Cref{fig:predicted-curves-edu}, we present the predicted surprisal derived from the EDU-scaled harmonic regression model. 
The predicted curves reflect the overall pattern of the observed surprisal, aligning particularly closely at unit boundaries.

We note that the differences between models are moderate, and their MSE still leaves considerable room for improvement. However, the HS hypothesis does not posit that periodic regularities stemming from discourse structure are the \emph{sole} determinants of surprisal contours; thus, a perfect fit for every individual surprisal value is not expected. 
Rather, our goal is to identify periodic regularities and investigate their connection to discourse structural elements.

\begin{table*}[ht!]
\centering
\small
\begin{tabular}{c@{\hspace{0.50em}}cc@{\hspace{0.50em}}cc@{\hspace{0.50em}}cc@{\hspace{0.50em}}c|c@{\hspace{0.50em}}cc@{\hspace{0.50em}}cc@{\hspace{0.50em}}cc@{\hspace{0.50em}}c}
\toprule
\multicolumn{8}{c}{\textbf{English}} & \multicolumn{8}{c}{\textbf{Spanish}} \\\cmidrule(lr){1-8} \cmidrule(lr){9-16}
\multicolumn{2}{c}{Document} & \multicolumn{2}{c}{EDU} & \multicolumn{2}{c}{Sentence} & \multicolumn{2}{c}{Paragraph} & \multicolumn{2}{c}{Document} & \multicolumn{2}{c}{EDU} & \multicolumn{2}{c}{Sentence} & \multicolumn{2}{c}{Paragraph} \\
$\component$ & $\amp_\component$ & $\component$ & $\amp_\component$ & $\component$ & $\amp_\component$ & $\component$ & $\amp_\component$ & $\component$ & $\amp_\component$ & $\component$ & $\amp_\component$ & $\component$ & $\amp_\component$ & $\component$ & $\amp_\component$ \\ \cmidrule(lr){1-2} \cmidrule(lr){3-4} \cmidrule(lr){5-6} \cmidrule(lr){7-8} \cmidrule(lr){9-10} \cmidrule(lr){11-12} \cmidrule(lr){13-14} \cmidrule(lr){15-16}
1 & 0.235$_{\texttt{10}}$ & 1 & 0.370$_{\texttt{10}}$ & 4 & 0.171$_{\texttt{10}}$ & 9 & 0.037$_{\texttt{10}}$ & 1 & 0.422$_{\texttt{10}}$ & 1 & 0.364$_{\texttt{10}}$ &  &  & 126 & 0.058$_{\texttt{7}}$ \\
 &  & 2 & 0.330$_{\texttt{10}}$ & 5 & 0.151$_{\texttt{10}}$ &  &  & 4 & 0.323$_{\texttt{10}}$ & 2 & 0.313$_{\texttt{10}}$ &  &  & 535 & 0.054$_{\texttt{10}}$ \\
 &  & 3 & 0.241$_{\texttt{10}}$ & 10 & 0.144$_{\texttt{10}}$ &  &  & 5 & 0.293$_{\texttt{10}}$ & 4 & 0.264$_{\texttt{10}}$ &  &  & 150 & 0.053$_{\texttt{8}}$ \\
\midrule
\midrule
\multicolumn{8}{c}{\textbf{German}} & \multicolumn{8}{c}{\textbf{Dutch}} \\\cmidrule(lr){1-8} \cmidrule(lr){9-16}
\multicolumn{2}{c}{Document} & \multicolumn{2}{c}{EDU} & \multicolumn{2}{c}{Sentence} & \multicolumn{2}{c}{Paragraph} & \multicolumn{2}{c}{Document} & \multicolumn{2}{c}{EDU} & \multicolumn{2}{c}{Sentence} & \multicolumn{2}{c}{Paragraph} \\
$\component$ & $\amp_\component$ & $\component$ & $\amp_\component$ & $\component$ & $\amp_\component$ & $\component$ & $\amp_\component$ & $\component$ & $\amp_\component$ & $\component$ & $\amp_\component$ & $\component$ & $\amp_\component$ & $\component$ & $\amp_\component$ \\ \cmidrule(lr){1-2} \cmidrule(lr){3-4} \cmidrule(lr){5-6} \cmidrule(lr){7-8} \cmidrule(lr){9-10} \cmidrule(lr){11-12} \cmidrule(lr){13-14} \cmidrule(lr){15-16}
4 & 0.165$_{\texttt{10}}$ & 1 & 0.599$_{\texttt{10}}$ & 10 & 0.101$_{\texttt{10}}$ & 11 & 0.087$_{\texttt{10}}$ & 5 & 0.153$_{\texttt{10}}$ & 1 & 0.470$_{\texttt{10}}$ & 3 & 0.198$_{\texttt{10}}$ & 10 & 0.095$_{\texttt{10}}$ \\
5 & 0.140$_{\texttt{10}}$ & 2 & 0.515$_{\texttt{10}}$ & 12 & 0.063$_{\texttt{10}}$ & 56 & 0.077$_{\texttt{10}}$ & 6 & 0.135$_{\texttt{10}}$ & 2 & 0.308$_{\texttt{10}}$ & 1 & 0.141$_{\texttt{10}}$ & 66 & 0.089$_{\texttt{10}}$ \\
6 & 0.137$_{\texttt{10}}$ & 3 & 0.380$_{\texttt{10}}$ & 13 & 0.059$_{\texttt{10}}$ & 345 & 0.066$_{\texttt{9}}$ & 7 & 0.103$_{\texttt{6}}$ & 3 & 0.247$_{\texttt{10}}$ &  &  & 8 & 0.074$_{\texttt{10}}$ \\
\midrule
\midrule
\multicolumn{8}{c}{\textbf{Basque}} & \multicolumn{8}{c}{\textbf{Brazilian Portuguese}} \\\cmidrule(lr){1-8} \cmidrule(lr){9-16}
\multicolumn{2}{c}{Document} & \multicolumn{2}{c}{EDU} & \multicolumn{2}{c}{Sentence} & \multicolumn{2}{c}{Paragraph} & \multicolumn{2}{c}{Document} & \multicolumn{2}{c}{EDU} & \multicolumn{2}{c}{Sentence} & \multicolumn{2}{c}{Paragraph} \\
$\component$ & $\amp_\component$ & $\component$ & $\amp_\component$ & $\component$ & $\amp_\component$ & $\component$ & $\amp_\component$ & $\component$ & $\amp_\component$ & $\component$ & $\amp_\component$ & $\component$ & $\amp_\component$ & $\component$ & $\amp_\component$ \\ \cmidrule(lr){1-2} \cmidrule(lr){3-4} \cmidrule(lr){5-6} \cmidrule(lr){7-8} \cmidrule(lr){9-10} \cmidrule(lr){11-12} \cmidrule(lr){13-14} \cmidrule(lr){15-16}
7 & 0.099$_{\texttt{10}}$ & 1 & 0.260$_{\texttt{10}}$ & 189 & 0.053$_{\texttt{0}}$ & 25 & 0.066$_{\texttt{10}}$ & 24 & 0.091$_{\texttt{10}}$ & 2 & 0.389$_{\texttt{10}}$ & 5 & 0.129$_{\texttt{10}}$ & 27 & 0.046$_{\texttt{10}}$ \\
6 & 0.098$_{\texttt{10}}$ & 2 & 0.196$_{\texttt{10}}$ & 112 & 0.043$_{\texttt{10}}$ & 651 & 0.054$_{\texttt{9}}$ & 14 & 0.065$_{\texttt{4}}$ & 3 & 0.312$_{\texttt{10}}$ & 4 & 0.084$_{\texttt{10}}$ & 34 & 0.041$_{\texttt{10}}$ \\
8 & 0.093$_{\texttt{10}}$ & 5 & 0.122$_{\texttt{10}}$ &  &  & 119 & 0.041$_{\texttt{7}}$ & 20 & 0.059$_{\texttt{3}}$ & 4 & 0.274$_{\texttt{10}}$ & 3 & 0.082$_{\texttt{10}}$ &  &  \\
\bottomrule
\end{tabular}
    \caption{Mean amplitudes ($\amp_\component$) of the three most dominant sinusoids that persist through feature selection in all ten cross-validation folds.  Subscripts indicate the number of folds in which each sinusoid is also statistically significant according to the ANOVA test (see \Cref{ssec:regularization-baseline}). Mean amplitudes for additional harmonic orders are reported in \Cref{tab:amplitudes-app}. 
    Fewer than three values indicate that fewer than three sinusoids persisted through feature selection across all folds.
    }
    \label{tab:amplitudes}
   \vspace{-10pt}
\end{table*}

\subsection{Contribution of Individual Sinusoids}
\label{ssec:individual-sinusoids}
To identify which periods most influence the shape of surprisal contours, we analyze the amplitudes estimated by the maximal harmonic regression model.
These amplitudes, which reflect the strength of each frequency (see \Cref{sec:harmonic_regression}), highlight the contributions of different harmonic components across structure types.
\Cref{tab:amplitudes} presents the mean amplitudes of the most dominant sinusoids, averaged across all cross-validation folds.
Subscripts denote the number of cross-validation folds in which a sinusoid is significant according to the ANOVA test ($\significance < 0.001$; see \Cref{ssec:regularization-baseline} and \Cref{sec:baseline-details} for more details).\looseness-1

EDU-scaled sinusoids, particularly those with lower orders ($\component \in [1,2,3,4]$), show the highest amplitudes in all languages except Spanish, where they rank second after document-scaled ones. 
The results for sentence-scaled sinusoids are mixed, while scaling by paragraph length leads to lower amplitudes compared to document scaling.
Notably, all EDU-scaled sinusoids that remain after feature selection are consistently significant across folds (see also \Cref{tab:amplitudes-app} for additional harmonics), indicating reliable predictive strength---an effect not seen for any other structure type.
Overall, our results reveal periodicity in surprisal contours, particularly at the EDU level. 
This yields evidence that EDUs play an important role in determining the information structure of discourse, corroborating results in \cite{tsipidi-etal-2024-surprise} while refining the form of the functional relationship between a unit's information and its position.\looseness-1

\section{Surprisal at Discourse Unit Boundaries}
\label{sec:surprisal-boundaries}
So far, we have observed significant periodicity in surprisal contours at the EDU level, with predicted harmonic curves closely aligning with discourse boundaries---points where surprisal tends to peak.

Moreover, boundary features emerge as the strongest individual predictors of surprisal, exhibiting the highest coefficients in both our baseline and maximal linear models (see \Cref{tab:coeff-bl-maximal-model} and \Cref{tab:baseline-features}).
These findings motivate a closer examination of the relationship between surprisal peaks, discourse unit boundaries, and periodicity. 
Specifically, we aim to understand the extent to which the contribution of harmonic components to the shape of surprisal contours is explained by their alignment with boundary peaks, as opposed to reflecting additional structure in the distribution of information---the latter interpretation being supported by the significance of harmonic component effects even after accounting for baseline predictors (as shown in \Cref{ssec:individual-sinusoids}).

To investigate the relationship between surprisal peaks and unit boundaries, we calculate the mean surprisal of tokens within one- and two-token windows immediately \textit{before} and \textit{after} paragraph, sentence, and EDU boundaries, and compare these values to the mean surprisal of tokens located farther from any boundary (i.e., all other tokens). 
We find that tokens preceding boundaries exhibit lower mean surprisal compared to non-boundary tokens, while those following boundaries considerably exhibit higher surprisal (see \Cref{tab:surprisal-boundaries-app} for a comparison of surprisal values before, after, and away from boundaries).
For example, in German, the single token before a paragraph boundary has a mean surprisal of $1.47 \pm 2.69$, while the one after has $7.75 \pm 5.31$. At sentence and EDU boundaries, surprisal shifts from $1.07 \pm 1.77$ to $7.06 \pm 3.73$ and from $1.30 \pm 1.87$ to $6.39 \pm 3.79$, respectively. This overall trend holds across languages and boundary types.\looseness=-1

\begin{figure}
\centering
    \includegraphics[width=\linewidth]{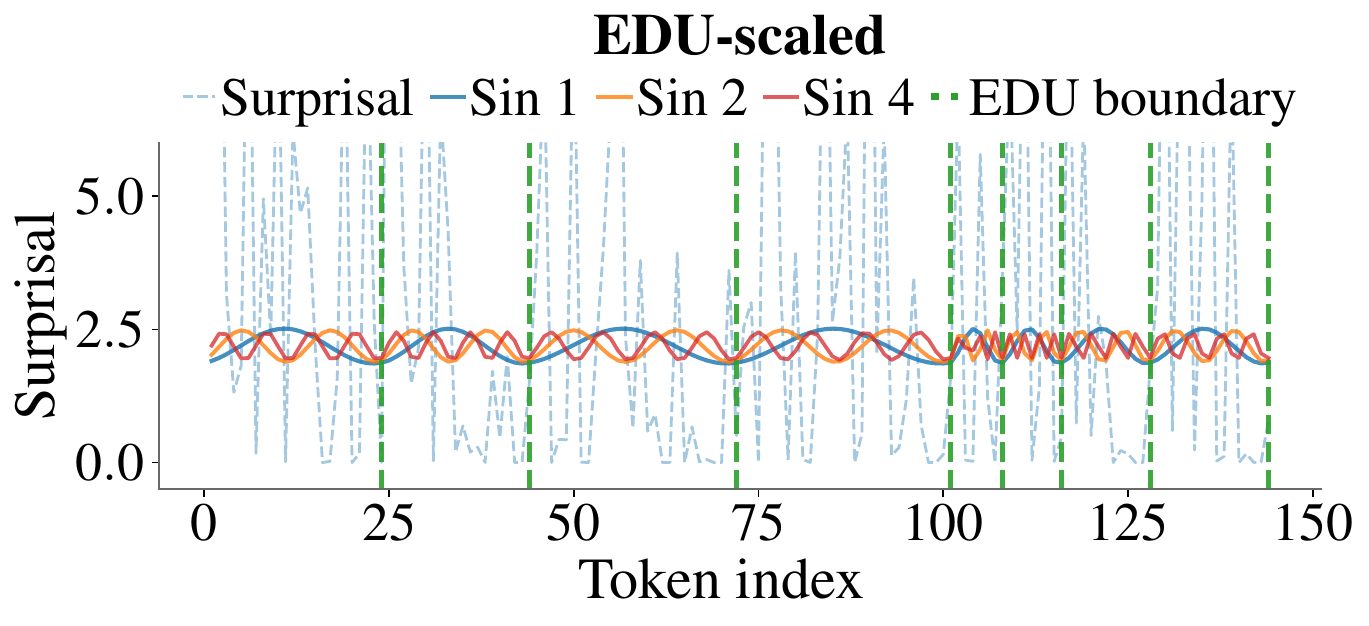}
    \caption{\textbf{Harmonic Structure in Surprisal Contours}. Top three most dominant sinusoids (EDU-scaled) in the maximal model for a Spanish text (doc \texttt{ma00059a}). Amplitudes signify the contribution to the overall variation, with higher amplitudes indicating a larger effect.\looseness-1
    }
    \label{fig:predictive-harmonics-vis-ma_00059a}
    \vspace{-15pt}
\end{figure}

To explore how this relates to periodicity, we focus on EDU boundaries, since they are the most predictive structure and subsume both sentence and paragraph boundaries. 
We visualize surprisal contours alongside unit boundaries and the highest-amplitude sinusoids. 
\Cref{fig:predictive-harmonics-vis-ma_00059a} shows this for a Spanish document, with further examples for other languages in \Cref{sec:sin-vis}. 
These visualizations reveal that the most prominent EDU-scaled sinusoids tend to intersect discourse boundaries at their troughs---points where the curve declines prior to the boundary and rises immediately after.
Taken together with the results presented in \Cref{sec:empirical-findings}, our findings suggest that information is systematically modulated around discourse boundaries---decreasing before and increasing after---and that surprisal exhibits periodicity which not only reflects this modulation but also extends beyond the immediate influence of boundary proximity.

These findings naturally give rise to further questions about the underlying mechanisms: Why is information organized to peak at discourse boundaries? And why is EDU-specific periodicity particularly prominent?
We can offer some preliminary speculations. 
Prior work on UID assumes a constant channel capacity, but there is substantial evidence that processing effort increases toward the end of syntactic phrases---due to greater memory demands and integration costs \citep{just1980theory, gibson1998locality, gibson2000dependency, Rayner01112000}---as well as toward the end of narrative events \citep{speer-zacks-2005, radvansky2010reading}. 
These findings imply that channel capacity decreases at the end of units and increases at their beginning. An optimal speaker would therefore modulate the information rate across transitions. 
Furthermore, context informativity tends to drop at the beginning of new structural units, as these often introduce new referents, topics, or discourse relations \citep{genzel-charniak-2002-entropy}. This, too, would result in increased surprisal around boundaries.
Nevertheless, the presence of numerous higher-frequency sinusoids with comparatively high amplitudes suggests the existence of meaningful structure at sub-EDU levels.
Investigating the role of these finer-grained units is a promising direction for future research.

\section{Conclusion}
\label{sec:conclusion}
We proposed a refinement of two hypotheses that aim to explain the structure of global information contours: the established Uniform Information Density hypothesis, which is agnostic to the nature and granularity of the units and structures being analyzed, and the more recent Structured Context hypothesis, which links per-unit surprisal rates to the unit’s position within the discourse structure.
The Harmonic Surprisal hypothesis gives more color to the phenomena that are not explained by the UID hypothesis, which remains neutral on why surprisal deviates from the mean (or the upper bound of channel capacity). It also makes stronger predictions than the SC hypothesis by introducing a periodic functional relationship between surprisal and discourse constituents. 

Our harmonic regression analysis of surprisal contours across six languages reveals clear periodic patterns, especially under EDU-based time scaling. Within discourse units, information is not evenly distributed: surprisal reliably decreases before and increases after discourse boundaries.
Although our focus is on global discourse-level trends, we also detect significant higher-order components in surprisal contours, indicating the presence of periodicity at smaller scales---such as syntactic units, words, with their subword tokens.
Future work could apply our time-scaled harmonic regression framework to these finer-grained structures to investigate their potential role in shaping surprisal dynamics.\looseness=-1

In conclusion, this contribution furthers the broader program of understanding the processing, aesthetic, and conventional constraints that shape the linguistic exchange of information.

\section*{Limitations}
One limitation of our study is the diversity of languages on which we test our hypotheses. 
Although Basque offers a typologically distinct language with a subject-object-verb (SOV) word order, the remaining languages are Indo-European, with a subject-verb-object (SVO) order. Future work could broaden the scope by including a more diverse set of languages.
Moreover, as already acknowledged in \Cref{sec:conclusion}, while our methodology is, in principle, applicable to structural elements at any scale, this study focused specifically on discourse-level constituents: EDUs, sentences, and paragraphs. 
Future work could extend our investigation by applying time-scaled harmonic regression to smaller linguistic structures---such as syntactic constructions, collocations, or even individual words---to explore whether similar periodic patterns emerge at those levels. 
Similarly, surprisal contours could be examined using base units of varying granularity, ranging from coarser units like full clauses or sentences to more fine-grained levels such as individual characters.

\section*{Ethics Statement}
We foresee no ethical problems with our work.

\section*{Acknowledgements}
We would like to thank Chlo{\'e} Braud for putting us in touch with \citet{vliet-etal-2011-building} and facilitating access to the Dutch RST corpus.
Mario Giulianelli was supported by an ETH Zürich Postdoctoral Fellowship.\looseness-1

\bibliography{anthology,custom}

\begin{thebibliography}{83}
\expandafter\ifx\csname natexlab\endcsname\relax\def\natexlab#1{#1}\fi

\bibitem[{Aurnhammer and Frank(2019)}]{aurnhammer2019evaluating}
Christoph Aurnhammer and Stefan~L. Frank. 2019.
\newblock \href
  {https://doi.org/https://doi.org/10.1016/j.neuropsychologia.2019.107198}
  {Evaluating information-theoretic measures of word prediction in naturalistic
  sentence reading}.
\newblock \emph{Neuropsychologia}, 134:107198.

\bibitem[{Aylett and Turk(2004)}]{aylett-turk-2004}
Matthew Aylett and Alice Turk. 2004.
\newblock \href {https://doi.org/10.1177/00238309040470010201} {The smooth
  signal redundancy hypothesis: A functional explanation for relationships
  between redundancy, prosodic prominence, and duration in spontaneous speech}.
\newblock \emph{Language and Speech}, 47(1):31--56.
\newblock PMID: 15298329.

\bibitem[{Aylett and Turk(2006)}]{aylett2006language}
Matthew Aylett and Alice Turk. 2006.
\newblock \href {https://doi.org/10.1121/1.2188331} {Language redundancy
  predicts syllabic duration and the spectral characteristics of vocalic
  syllable nuclei}.
\newblock \emph{The Journal of the Acoustical Society of America},
  119(5):3048--3058.

\bibitem[{Aylett(1999)}]{aylett1999}
{Matthew P.} Aylett. 1999.
\newblock Stochastic suprasegmentals: Relationships between redundancy,
  prosodic structure and syllabic duration.
\newblock In \emph{Proceedings of the XIVth International Congress of Phonetic
  Sciences}, pages 289--292. American Institute of Physics.

\bibitem[{Bell et~al.(2003)Bell, Jurafsky, Fosler{-}Lussier, Girand, Gregory,
  and Gildea}]{bell2003effects}
Alan Bell, Daniel Jurafsky, Eric Fosler{-}Lussier, Cynthia Girand, Michelle
  Gregory, and Daniel Gildea. 2003.
\newblock \href {https://doi.org/10.1121/1.1534836} {Effects of disfluencies,
  predictability, and utterance position on word form variation in {E}nglish
  conversation}.
\newblock \emph{The Journal of the Acoustical Society of America},
  113(2):1001--1024.

\bibitem[{Bergey and DeDeo(2024)}]{bergey2024producing}
Claire~Augusta Bergey and Simon DeDeo. 2024.
\newblock \href {http://arxiv.org/abs/2403.08890} {From "um" to "yeah":
  Producing, predicting, and regulating information flow in human
  conversation}.

\bibitem[{Cardoso et~al.(2011)Cardoso, Maziero, Jorge, Seno, Felippo, Rino, das
  Graças Volpe~Nunes, and Pardo}]{Cardoso-etal-2011}
Paula C.~F. Cardoso, Erick~G. Maziero, Mara Luca~Castro Jorge, Eloize R.~M.
  Seno, Ariani~Di Felippo, Lucia Helena~Machado Rino, Maria das Graças
  Volpe~Nunes, and Thiago A.~S. Pardo. 2011.
\newblock \href {https://sites.icmc.usp.br/taspardo/RST2011-CardosoEtAl1.pdf}
  {Cstnews - a discourse-annotated corpus for single and multidocument
  summarization of news texts in brazilian portuguese}.
\newblock In \emph{Proceedings of the 3rd RST Brazilian Meeting}, pages
  88--105.

\bibitem[{Carlson and Marcu(2001)}]{carlson-marcu-2001-discourse}
Lynn Carlson and Daniel Marcu. 2001.
\newblock \href
  {https://web.archive.org/web/20170808131213id_/https://www.isi.edu/~marcu/discourse/tagging-ref-manual.pdf}
  {Discourse tagging reference manual}.
\newblock Technical Report ISI-TR-545, University of Southern California
  Information Sciences Institute.

\bibitem[{Carlson et~al.(2001)Carlson, Marcu, and
  Okurovsky}]{carlson-etal-2001-building}
Lynn Carlson, Daniel Marcu, and Mary~Ellen Okurovsky. 2001.
\newblock \href {https://aclanthology.org/W01-1605} {Building a
  discourse-tagged corpus in the framework of {R}hetorical {S}tructure
  {T}heory}.
\newblock In \emph{Proceedings of the Second {SIG}dial Workshop on Discourse
  and Dialogue}.

\bibitem[{Cervantes and Gainer(1992)}]{cervantes1992effects}
Raoul Cervantes and Glenn Gainer. 1992.
\newblock \href {http://www.jstor.org/stable/3586886} {The effects of syntactic
  simplification and repetition on listening comprehension}.
\newblock \emph{TESOL Quarterly}, 26(4):767--770.
\newblock Wiley.

\bibitem[{Clark and Wilkes-Gibbs(1986)}]{clark-1986-referring}
Herbert~H. Clark and Deanna Wilkes-Gibbs. 1986.
\newblock \href {https://doi.org/https://doi.org/10.1016/0010-0277(86)90010-7}
  {Referring as a collaborative process}.
\newblock \emph{Cognition}, 22(1):1--39.

\bibitem[{Cohen~Priva(2015)}]{cohenpriva2015informativity}
Uriel Cohen~Priva. 2015.
\newblock \href {https://doi.org/10.1515/lp-2015-0008} {Informativity affects
  consonant duration and deletion rates}.
\newblock \emph{Laboratory Phonology}, 6(2).

\bibitem[{Collins(2014)}]{collins2014information}
Michael~Xavier Collins. 2014.
\newblock \href {https://doi.org/https://doi.org/10.1007/s10936-013-9273-3}
  {Information density and dependency length as complementary cognitive
  models}.
\newblock \emph{Journal of Psycholinguistic Research}, 43:651--681.

\bibitem[{da~Cunha et~al.(2011)da~Cunha, Torres-Moreno, and
  Sierra}]{da-cunha-etal-2011-development}
Iria da~Cunha, Juan-Manuel Torres-Moreno, and Gerardo Sierra. 2011.
\newblock \href {https://aclanthology.org/W11-0401} {On the development of the
  {RST} {S}panish treebank}.
\newblock In \emph{Proceedings of the 5th Linguistic Annotation Workshop},
  pages 1--10, Portland, Oregon, USA. Association for Computational
  Linguistics.

\bibitem[{Elman(1990)}]{elman1990finding}
Jeffrey~L. Elman. 1990.
\newblock \href
  {https://www.sciencedirect.com/science/article/pii/036402139090002E} {Finding
  structure in time}.
\newblock \emph{Cognitive Science}, 14(2):179--211.
\newblock Wiley Online Library.

\bibitem[{Etxaniz et~al.(2024)Etxaniz, Sainz, Perez, Aldabe, Rigau, Agirre,
  Ormazabal, Artetxe, and Soroa}]{etxaniz2024latxa}
Julen Etxaniz, Oscar Sainz, Naiara Perez, Itziar Aldabe, German Rigau, Eneko
  Agirre, Aitor Ormazabal, Mikel Artetxe, and Aitor Soroa. 2024.
\newblock \href {http://arxiv.org/abs/2403.20266} {{L}atxa: An open language
  model and evaluation suite for {B}asque}.

\bibitem[{Fenk and Fenk(1980)}]{fenk1980konstanz}
August Fenk and Gertraud Fenk. 1980.
\newblock \href
  {https://wwwg.uni-klu.ac.at/mk0/personal/bedienst/Kurzzeitgedaechtnis1980.pdf}
  {{Konstanz im Kurzzeitged{\"a}chtnis-konstanz im sprachlichen
  Informationsflu{\ss}?}}
\newblock \emph{Zeitschrift f{\"u}r experimentelle und angewandte Psychologie},
  27(3):400--414.

\bibitem[{Fernandez~Monsalve et~al.(2012)Fernandez~Monsalve, Frank, and
  Vigliocco}]{fernandez-monsalve-etal-2012-lexical}
Irene Fernandez~Monsalve, Stefan~L. Frank, and Gabriella Vigliocco. 2012.
\newblock \href {https://aclanthology.org/E12-1041} {Lexical surprisal as a
  general predictor of reading time}.
\newblock In \emph{Proceedings of the 13th Conference of the {E}uropean Chapter
  of the Association for Computational Linguistics}, pages 398--408, Avignon,
  France. Association for Computational Linguistics.

\bibitem[{Frank and Jaeger(2008)}]{frank2008speaking}
Austin~F Frank and T~Florain Jaeger. 2008.
\newblock \href {https://escholarship.org/content/qt7d08h6j4/qt7d08h6j4.pdf}
  {Speaking rationally: Uniform information density as an optimal strategy for
  language production}.
\newblock In \emph{Proceedings of the annual meeting of the cognitive science
  society}, volume~30.

\bibitem[{Frank et~al.(2015)Frank, Otten, Galli, and Vigliocco}]{frank2015erp}
Stefan~L Frank, Leun~J Otten, Giulia Galli, and Gabriella Vigliocco. 2015.
\newblock \href {https://doi.org/10.1016/j.bandl.2014.10.006} {The {ERP}
  response to the amount of information conveyed by words in sentences}.
\newblock \emph{Brain and Language}, 140:1--11.

\bibitem[{Frohmann et~al.(2024)Frohmann, Sterner, Vuli{\'c}, Minixhofer, and
  Schedl}]{frohmann-etal-2024-segment}
Markus Frohmann, Igor Sterner, Ivan Vuli{\'c}, Benjamin Minixhofer, and Markus
  Schedl. 2024.
\newblock \href {https://doi.org/10.18653/v1/2024.emnlp-main.665} {Segment any
  text: A universal approach for robust, efficient and adaptable sentence
  segmentation}.
\newblock In \emph{Proceedings of the 2024 Conference on Empirical Methods in
  Natural Language Processing}, pages 11908--11941, Miami, Florida, USA.
  Association for Computational Linguistics.

\bibitem[{Futrell(2023)}]{futrell-2023-production}
Richard Futrell. 2023.
\newblock \href {https://doi.org/10.1073/pnas.2220593120}
  {Information-theoretic principles in incremental language production}.
\newblock \emph{Proceedings of the National Academy of Sciences},
  120(39):e2220593120.

\bibitem[{Genzel and Charniak(2002)}]{genzel-charniak-2002-entropy}
Dmitriy Genzel and Eugene Charniak. 2002.
\newblock \href {https://doi.org/10.3115/1073083.1073117} {Entropy rate
  constancy in text}.
\newblock In \emph{Proceedings of the 40th Annual Meeting of the Association
  for Computational Linguistics}, pages 199--206, Philadelphia, Pennsylvania,
  USA. Association for Computational Linguistics.

\bibitem[{Genzel and Charniak(2003)}]{genzel-charniak-2003-variation}
Dmitriy Genzel and Eugene Charniak. 2003.
\newblock \href {https://aclanthology.org/W03-1009} {Variation of entropy and
  parse trees of sentences as a function of the sentence number}.
\newblock In \emph{Proceedings of the 2003 Conference on Empirical Methods in
  Natural Language Processing}, pages 65--72.

\bibitem[{Gibson(1998)}]{gibson1998locality}
Edward Gibson. 1998.
\newblock \href {https://doi.org/https://doi.org/10.1016/S0010-0277(98)00034-1}
  {Linguistic complexity: locality of syntactic dependencies}.
\newblock \emph{Cognition}, 68(1):1--76.

\bibitem[{Gibson(2000)}]{gibson2000dependency}
Edward Gibson. 2000.
\newblock \href
  {https://tedlab.mit.edu/tedlab_website/researchpapers/Gibson_2000_DLT.pdf}
  {The dependency locality theory: A distance-based theory of linguistic
  complexity}.
\newblock \emph{Image, language, brain/MIT Press}.

\bibitem[{Gibson et~al.(2019)Gibson, Futrell, Piantadosi, Dautriche, Mahowald,
  Bergen, and Levy}]{gibson2019efficiency}
Edward Gibson, Richard Futrell, Steven~P. Piantadosi, Isabelle Dautriche, Kyle
  Mahowald, Leon Bergen, and Roger Levy. 2019.
\newblock \href {https://doi.org/https://doi.org/10.1016/j.tics.2019.02.003}
  {How efficiency shapes human language}.
\newblock \emph{Trends in Cognitive Sciences}, 23(5):389--407.

\bibitem[{Giulianelli and
  Fern{\'a}ndez(2021)}]{giulianelli-fernandez-2021-analysing}
Mario Giulianelli and Raquel Fern{\'a}ndez. 2021.
\newblock \href {https://doi.org/10.18653/v1/2021.conll-1.50} {Analysing human
  strategies of information transmission as a function of discourse context}.
\newblock In \emph{Proceedings of the 25th Conference on Computational Natural
  Language Learning}, pages 647--660, Online. Association for Computational
  Linguistics.

\bibitem[{Giulianelli et~al.(2024{\natexlab{a}})Giulianelli, Malagutti,
  Gastaldi, DuSell, Vieira, and Cotterell}]{giulianelli-etal-2024-proper}
Mario Giulianelli, Luca Malagutti, Juan~Luis Gastaldi, Brian DuSell, Tim
  Vieira, and Ryan Cotterell. 2024{\natexlab{a}}.
\newblock \href {https://doi.org/10.18653/v1/2024.emnlp-main.1032} {On the
  proper treatment of tokenization in psycholinguistics}.
\newblock In \emph{Proceedings of the 2024 Conference on Empirical Methods in
  Natural Language Processing}, pages 18556--18572, Miami, Florida, USA.
  Association for Computational Linguistics.

\bibitem[{Giulianelli et~al.(2024{\natexlab{b}})Giulianelli, Opedal, and
  Cotterell}]{giulianelli-etal-2024-generalized}
Mario Giulianelli, Andreas Opedal, and Ryan Cotterell. 2024{\natexlab{b}}.
\newblock \href {https://doi.org/10.18653/v1/2024.findings-emnlp.682}
  {Generalized measures of anticipation and responsivity in online language
  processing}.
\newblock In \emph{Findings of the Association for Computational Linguistics:
  EMNLP 2024}, pages 11648--11669, Miami, Florida, USA. Association for
  Computational Linguistics.

\bibitem[{Giulianelli et~al.(2021)Giulianelli, Sinclair, and
  Fern{\'a}ndez}]{giulianelli-etal-2021-information}
Mario Giulianelli, Arabella Sinclair, and Raquel Fern{\'a}ndez. 2021.
\newblock \href {https://doi.org/10.18653/v1/2021.emnlp-main.652} {Is
  information density uniform in task-oriented dialogues?}
\newblock In \emph{Proceedings of the 2021 Conference on Empirical Methods in
  Natural Language Processing}, pages 8271--8283, Online and Punta Cana,
  Dominican Republic. Association for Computational Linguistics.

\bibitem[{Giulianelli et~al.(2024{\natexlab{c}})Giulianelli, Wallbridge,
  Cotterell, and Fernández}]{giulianelli-etal-2024-incremental}
Mario Giulianelli, Sarenne Wallbridge, Ryan Cotterell, and Raquel Fernández.
  2024{\natexlab{c}}.
\newblock \href {https://doi.org/10.31234/osf.io/fhp84} {Incremental
  alternative sampling as a lens into the temporal and representational
  resolution of linguistic prediction}.

\bibitem[{Giulianelli et~al.(2023)Giulianelli, Wallbridge, and
  Fern{\'a}ndez}]{giulianelli-etal-2023-information}
Mario Giulianelli, Sarenne Wallbridge, and Raquel Fern{\'a}ndez. 2023.
\newblock \href {https://doi.org/10.18653/v1/2023.emnlp-main.343} {Information
  value: Measuring utterance predictability as distance from plausible
  alternatives}.
\newblock In \emph{Proceedings of the 2023 Conference on Empirical Methods in
  Natural Language Processing}, pages 5633--5653, Singapore. Association for
  Computational Linguistics.

\bibitem[{Goodkind and Bicknell(2018)}]{goodkind2018predictive}
Adam Goodkind and Klinton Bicknell. 2018.
\newblock \href {https://doi.org/10.18653/v1/W18-0102} {Predictive power of
  word surprisal for reading times is a linear function of language model
  quality}.
\newblock In \emph{Proceedings of the 8th Workshop on Cognitive Modeling and
  Computational Linguistics ({CMCL} 2018)}, pages 10--18, Salt Lake City, Utah.
  Association for Computational Linguistics.

\bibitem[{Holm(1979)}]{holm1979simple}
Sture Holm. 1979.
\newblock A simple sequentially rejective multiple test procedure.
\newblock \emph{Scandinavian journal of statistics}, pages 65--70.

\bibitem[{Huber et~al.(2024)Huber, Sauppe, Isasi-Isasmendi,
  Bornkessel-Schlesewsky, Merlo, and Bickel}]{Huber2024-vt}
Eva Huber, Sebastian Sauppe, Arrate Isasi-Isasmendi, Ina
  Bornkessel-Schlesewsky, Paola Merlo, and Balthasar Bickel. 2024.
\newblock \href {https://doi.org/10.1162/nol_a_00121} {Surprisal from language
  models can predict {ERP}s in processing predicate-argument structures only if
  enriched by an agent preference principle}.
\newblock \emph{Neurobiology of Language}, 5(1):167--200.

\bibitem[{Iruskieta et~al.(2013)Iruskieta, Aranzabe, de~Ilarraza,
  Gonzalez-Dios, Lersundi, and de~la Calle}]{iruskieta-etal-2013-RSTBasque}
Mikel Iruskieta, Mar\'{i}a~J. Aranzabe, Arantza~Diaz de~Ilarraza, Itziar
  Gonzalez-Dios, Mikel Lersundi, and Oier~Lopez de~la Calle. 2013.
\newblock \href
  {https://www.ixa.eus/sites/default/files/dokumentuak/3960/2013RST-Basque-TB.pdf}
  {The rst basque treebank: An online search interface to check rhetorical
  relations}.
\newblock In \emph{Proceedings of the 4th Workshop on RST and Discourse
  Studies}.

\bibitem[{Jaeger(2010)}]{jaeger2010redundancy}
T.~Florian Jaeger. 2010.
\newblock \href
  {https://doi.org/https://doi.org/10.1016/j.cogpsych.2010.02.002} {Redundancy
  and reduction: Speakers manage syntactic information density}.
\newblock \emph{Cognitive Psychology}, 61(1):23--62.

\bibitem[{Just and Carpenter(1980)}]{just1980theory}
Marcel~A Just and Patricia~A Carpenter. 1980.
\newblock \href
  {https://psycnet.apa.org/doiLanding?doi=10.1037%2F0033-295X.87.4.329} {A
  theory of reading: from eye fixations to comprehension.}
\newblock \emph{Psychological review}, 87(4):329.

\bibitem[{Keith~Rayner and Duffy(2000)}]{Rayner01112000}
Gretchen~Kambe Keith~Rayner and Susan~A. Duffy. 2000.
\newblock \href {https://doi.org/10.1080/713755934} {The effect of clause
  wrap-up on eye movements during reading}.
\newblock \emph{The Quarterly Journal of Experimental Psychology Section A},
  53(4):1061--1080.
\newblock PMID: 11131813.

\bibitem[{Keller(2004)}]{keller-2004-entropy}
Frank Keller. 2004.
\newblock \href {https://aclanthology.org/W04-3241} {The entropy rate principle
  as a predictor of processing effort: An evaluation against eye-tracking
  data}.
\newblock In \emph{Proceedings of the 2004 Conference on Empirical Methods in
  Natural Language Processing}, pages 317--324, Barcelona, Spain. Association
  for Computational Linguistics.

\bibitem[{Levy and Jaeger(2006)}]{levy-jaeger-2007-reduction}
Roger Levy and T.~Florian Jaeger. 2006.
\newblock \href
  {https://proceedings.neurips.cc/paper_files/paper/2006/file/c6a01432c8138d46ba39957a8250e027-Paper.pdf}
  {Speakers optimize information density through syntactic reduction}.
\newblock In \emph{Advances in Neural Information Processing Systems},
  volume~19. MIT Press.

\bibitem[{Lewis(1894)}]{lewis1894history}
Edwin~Herbert Lewis. 1894.
\newblock \href {https://archive.org/details/historyenglishp00lewigoog}
  {\emph{The history of the English paragraph}}.
\newblock University of Chicago Press.

\bibitem[{Li and Futrell(2024)}]{li2024information}
Jiaxuan Li and Richard Futrell. 2024.
\newblock \href
  {https://escholarship.org/content/qt1fd682nd/qt1fd682nd_noSplash_e733707513b8f4be3d407d3f029acd2b.pdf?t=sgric3}
  {An information-theoretic model of shallow and deep language comprehension}.
\newblock In \emph{Proceedings of the Annual Meeting of the Cognitive Science
  Society}, volume~46.

\bibitem[{Ma{\"e}s et~al.(2022)Ma{\"e}s, Blache, and
  Becerra-Bonache}]{maes2022shared}
Eliot Ma{\"e}s, Philippe Blache, and Leonor Becerra-Bonache. 2022.
\newblock \href {https://hal.science/hal-04151675/document} {Shared knowledge
  in natural conversations: {c}an entropy metrics shed light on information
  transfers?}
\newblock In \emph{26th Conference on Computational Natural Language Learning},
  pages 213--227.

\bibitem[{Mahowald et~al.(2013)Mahowald, Fedorenko, Piantadosi, and
  Gibson}]{maholwald-2013-info}
Kyle Mahowald, Evelina Fedorenko, Steven~T. Piantadosi, and Edward Gibson.
  2013.
\newblock \href
  {https://doi.org/https://doi.org/10.1016/j.cognition.2012.09.010}
  {Info/information theory: Speakers choose shorter words in predictive
  contexts}.
\newblock \emph{Cognition}, 126(2):313--318.

\bibitem[{Mann and Thompson(1988)}]{mann-thompson-rst-1988}
William~C. Mann and Sandra~A. Thompson. 1988.
\newblock \href {https://doi.org/doi:10.1515/text.1.1988.8.3.243} {Rhetorical
  structure theory: Toward a functional theory of text organization}.
\newblock \emph{Text - Interdisciplinary Journal for the Study of Discourse},
  8(3):243--281.

\bibitem[{Marcus et~al.(1999)Marcus, Santorini, Marcinkiewicz, and
  Taylor}]{marcus1999treebank}
Mitchell~P. Marcus, Beatrice Santorini, Mary~Ann Marcinkiewicz, and Ann Taylor.
  1999.
\newblock \href {https://catalog.ldc.upenn.edu/LDC99T42} {Treebank-3}.
\newblock \emph{Linguistic Data Consortium}, 14.

\bibitem[{Meister et~al.(2024)Meister, Giulianelli, and
  Pimentel}]{meister-etal-2024-towards}
Clara Meister, Mario Giulianelli, and Tiago Pimentel. 2024.
\newblock \href {https://doi.org/10.18653/v1/2024.emnlp-main.921} {Towards a
  similarity-adjusted surprisal theory}.
\newblock In \emph{Proceedings of the 2024 Conference on Empirical Methods in
  Natural Language Processing}, pages 16485--16498, Miami, Florida, USA.
  Association for Computational Linguistics.

\bibitem[{Meister et~al.(2021)Meister, Pimentel, Haller, J{\"a}ger, Cotterell,
  and Levy}]{meister-etal-2021-revisiting}
Clara Meister, Tiago Pimentel, Patrick Haller, Lena J{\"a}ger, Ryan Cotterell,
  and Roger Levy. 2021.
\newblock \href {https://doi.org/10.18653/v1/2021.emnlp-main.74} {Revisiting
  the {U}niform {I}nformation {D}ensity hypothesis}.
\newblock In \emph{Proceedings of the 2021 Conference on Empirical Methods in
  Natural Language Processing}, pages 963--980, Online and Punta Cana,
  Dominican Republic. Association for Computational Linguistics.

\bibitem[{Minixhofer et~al.(2023)Minixhofer, Pfeiffer, and
  Vuli{\'c}}]{minixhofer-etal-2023-wheres}
Benjamin Minixhofer, Jonas Pfeiffer, and Ivan Vuli{\'c}. 2023.
\newblock \href {https://aclanthology.org/2023.acl-long.398} {Where{'}s the
  point? self-supervised multilingual punctuation-agnostic sentence
  segmentation}.
\newblock In \emph{Proceedings of the 61st Annual Meeting of the Association
  for Computational Linguistics (Volume 1: Long Papers)}, pages 7215--7235,
  Toronto, Canada. Association for Computational Linguistics.

\bibitem[{Peng et~al.(2024)Peng, Quesnelle, Fan, and Shippole}]{peng2024yarn}
Bowen Peng, Jeffrey Quesnelle, Honglu Fan, and Enrico Shippole. 2024.
\newblock \href {https://openreview.net/forum?id=wHBfxhZu1u} {Ya{RN}:
  {Efficient} context window extension of large language models}.
\newblock In \emph{The Twelfth International Conference on Learning
  Representations}.

\bibitem[{Pires et~al.(2023)Pires, Abonizio, Almeida, and
  Nogueira}]{pires2023sabia}
Ramon Pires, Hugo Abonizio, Thales~Sales Almeida, and Rodrigo Nogueira. 2023.
\newblock \href {https://doi.org/10.1007/978-3-031-45392-2_15} {Sabi{\'a}:
  Portuguese large language models}.
\newblock In \emph{Intelligent Systems}, pages 226--240, Cham. Springer Nature
  Switzerland.

\bibitem[{Qian and Jaeger(2011)}]{qian2011topic}
Ting Qian and T.~Florian Jaeger. 2011.
\newblock \href {https://escholarship.org/content/qt6b0712jg/qt6b0712jg.pdf}
  {Topic shift in efficient discourse production}.
\newblock In \emph{Proceedings of the Annual Meeting of the Cognitive Science
  Society}, volume~33.

\bibitem[{Rabovsky et~al.(2018)Rabovsky, Hansen, and
  McClelland}]{rabovsky2018modelling}
Milena Rabovsky, Steven~S Hansen, and James~L McClelland. 2018.
\newblock \href
  {https://doi.org/https://www.nature.com/articles/s41562-018-0406-4}
  {Modelling the {N}400 brain potential as change in a probabilistic
  representation of meaning}.
\newblock \emph{Nature Human Behaviour}, 2(9):693--705.

\bibitem[{Radvansky and Copeland(2010)}]{radvansky2010reading}
Gabriel~A. Radvansky and David~E. Copeland. 2010.
\newblock \href {https://doi.org/10.1037/a0017258} {Reading times and the
  detection of event shift processing.}
\newblock \emph{Journal of Experimental Psychology: Learning, Memory, and
  Cognition}, 36(1):210--216.
\newblock Place: US Publisher: American Psychological Association.

\bibitem[{Redeker et~al.(2012)Redeker, Berzl{\'a}novich, van~der Vliet, Bouma,
  and Egg}]{redeker-etal-2012-multi}
Gisela Redeker, Ildik{\'o} Berzl{\'a}novich, Nynke van~der Vliet, Gosse Bouma,
  and Markus Egg. 2012.
\newblock \href
  {http://www.lrec-conf.org/proceedings/lrec2012/pdf/887_Paper.pdf}
  {Multi-layer discourse annotation of a {D}utch text corpus}.
\newblock In \emph{Proceedings of the Eighth International Conference on
  Language Resources and Evaluation ({LREC}'12)}, pages 2820--2825, Istanbul,
  Turkey. European Language Resources Association (ELRA).

\bibitem[{Schrimpf et~al.(2021)Schrimpf, Blank, Tuckute, Kauf, Hosseini,
  Kanwisher, Tenenbaum, and Fedorenko}]{schrimpf2021}
Martin Schrimpf, Idan~Asher Blank, Greta Tuckute, Carina Kauf, Eghbal~A.
  Hosseini, Nancy Kanwisher, Joshua~B. Tenenbaum, and Evelina Fedorenko. 2021.
\newblock \href {https://doi.org/10.1073/pnas.2105646118} {The neural
  architecture of language: Integrative modeling converges on predictive
  processing}.
\newblock \emph{Proceedings of the National Academy of Sciences}, 118(45).

\bibitem[{Shain et~al.(2020)Shain, Blank, {van Schijndel}, Schuler, and
  Fedorenko}]{shain2020fmri}
Cory Shain, Idan~Asher Blank, Marten {van Schijndel}, William Schuler, and
  Evelina Fedorenko. 2020.
\newblock \href
  {https://doi.org/https://doi.org/10.1016/j.neuropsychologia.2019.107307}
  {fmri reveals language-specific predictive coding during naturalistic
  sentence comprehension}.
\newblock \emph{Neuropsychologia}, 138:107307.

\bibitem[{Shain et~al.(2024)Shain, Meister, Pimentel, Cotterell, and
  Levy}]{shainetal24}
Cory Shain, Clara Meister, Tiago Pimentel, Ryan Cotterell, and Roger Levy.
  2024.
\newblock \href {https://doi.org/https://doi.org/10.1073/pnas.2307876121}
  {Large-scale evidence for logarithmic effects of word predictability on
  reading time}.
\newblock \emph{Proceedings of the National Academy of Sciences},
  121(10):e2307876121.

\bibitem[{Shannon(1948)}]{shannon1948mathematical}
Claude~E. Shannon. 1948.
\newblock \href {https://ieeexplore.ieee.org/document/6773024} {A mathematical
  theory of communication}.
\newblock \emph{The Bell System Technical Journal}, 27(3):379--423.

\bibitem[{{S}kipper {S}eabold and {J}osef
  {P}erktold(2010)}]{seabold-proc-scipy-2010}
{S}kipper {S}eabold and {J}osef {P}erktold. 2010.
\newblock \href {https://doi.org/10.25080/Majora-92bf1922-011} {{S}tatsmodels:
  {E}conometric and {S}tatistical {M}odeling with {P}ython}.
\newblock In \emph{{P}roceedings of the 9th {P}ython in {S}cience
  {C}onference}, pages 92 -- 96.

\bibitem[{Slaats et~al.(2024)Slaats, Meyer, and Martin}]{slaats2024lexical}
Sophie Slaats, Antje~S. Meyer, and Andrea~E. Martin. 2024.
\newblock \href {https://doi.org/10.1162/nol_a_00155} {Lexical surprisal shapes
  the time course of syntactic structure building}.
\newblock \emph{Neurobiology of Language}, 5(4):942--980.

\bibitem[{Smith and Levy(2013)}]{smith-levy-2013}
Nathaniel~J. Smith and Roger Levy. 2013.
\newblock \href
  {https://doi.org/https://doi.org/10.1016/j.cognition.2013.02.013} {The effect
  of word predictability on reading time is logarithmic}.
\newblock \emph{Cognition}, 128(3):302--319.

\bibitem[{Speer and Zacks(2005)}]{speer-zacks-2005}
Nicole~K. Speer and Jeffrey~M. Zacks. 2005.
\newblock \href {https://doi.org/https://doi.org/10.1016/j.jml.2005.02.009}
  {Temporal changes as event boundaries: Processing and memory consequences of
  narrative time shifts}.
\newblock \emph{Journal of Memory and Language}, 53(1):125--140.

\bibitem[{Stede(2004)}]{stede-2004-potsdam}
Manfred Stede. 2004.
\newblock \href {https://aclanthology.org/W04-0213} {The {P}otsdam commentary
  corpus}.
\newblock In \emph{Proceedings of the Workshop on Discourse Annotation}, pages
  96--102, Barcelona, Spain. Association for Computational Linguistics.

\bibitem[{Stede and Neumann(2014)}]{stede-neumann-2014-potsdam}
Manfred Stede and Arne Neumann. 2014.
\newblock \href
  {http://www.lrec-conf.org/proceedings/lrec2014/pdf/579_Paper.pdf} {{P}otsdam
  commentary corpus 2.0: Annotation for discourse research}.
\newblock In \emph{Proceedings of the Ninth International Conference on
  Language Resources and Evaluation ({LREC}'14)}, pages 925--929, Reykjavik,
  Iceland. European Language Resources Association (ELRA).

\bibitem[{Torabi~Asr and Demberg(2015)}]{torabi-asr-demberg-2015-uniform}
Fatemeh Torabi~Asr and Vera Demberg. 2015.
\newblock \href {https://aclanthology.org/W15-0117} {Uniform surprisal at the
  level of discourse relations: Negation markers and discourse connective
  omission}.
\newblock In \emph{Proceedings of the 11th International Conference on
  Computational Semantics}, pages 118--128, London, UK. Association for
  Computational Linguistics.

\bibitem[{Tsipidi et~al.(2024)Tsipidi, Nowak, Cotterell, Wilcox, Giulianelli,
  and Warstadt}]{tsipidi-etal-2024-surprise}
Eleftheria Tsipidi, Franz Nowak, Ryan Cotterell, Ethan Wilcox, Mario
  Giulianelli, and Alex Warstadt. 2024.
\newblock \href {https://doi.org/10.18653/v1/2024.emnlp-main.1047} {Surprise!
  {U}niform {I}nformation {D}ensity isn't the whole story: Predicting surprisal
  contours in long-form discourse}.
\newblock In \emph{Proceedings of the 2024 Conference on Empirical Methods in
  Natural Language Processing}, pages 18820--18836, Miami, Florida, USA.
  Association for Computational Linguistics.

\bibitem[{{van der Vliet} et~al.(2011){van der Vliet}, Berzl{\'a}novich, Bouma,
  Egg, and Redeker}]{vliet-etal-2011-building}
N.H. {van der Vliet}, I.~Berzl{\'a}novich, G.~Bouma, M.~Egg, and G.~Redeker.
  2011.
\newblock Building a discourse-annotated dutch text corpus.
\newblock In \emph{Beyond Semantics}, volume~3 of \emph{Bochumer Linguistische
  Arbeitsberichte}, pages 157 -- 171. Ruhr-Universit{\"a}t Bochum.
\newblock 2011/g.bouma/pub005, 2011/g.redeker/pub001 E-publication only.

\bibitem[{Vanroy(2024)}]{vanroy2024geitje7bultraconversational}
Bram Vanroy. 2024.
\newblock \href {http://arxiv.org/abs/2412.04092} {Geitje 7b ultra: A
  conversational model for dutch}.

\bibitem[{Verma et~al.(2023)Verma, Tomlin, and
  Klein}]{verma-etal-2023-revisiting}
Vivek Verma, Nicholas Tomlin, and Dan Klein. 2023.
\newblock \href {https://doi.org/10.18653/v1/2023.findings-emnlp.1039}
  {Revisiting entropy rate constancy in text}.
\newblock In \emph{Findings of the Association for Computational Linguistics:
  EMNLP 2023}, pages 15537--15549, Singapore.

\bibitem[{Vieira et~al.(2024)Vieira, LeBrun, Giulianelli, Gastaldi, DuSell,
  Terilla, O'Donnell, and Cotterell}]{vieira-2024-token-to-char-lm}
Tim Vieira, Ben LeBrun, Mario Giulianelli, Juan~Luis Gastaldi, Brian DuSell,
  John Terilla, Timothy~J. O'Donnell, and Ryan Cotterell. 2024.
\newblock \href {http://arxiv.org/abs/2412.03719} {From language models over
  tokens to language models over characters}.

\bibitem[{Wallbridge et~al.(2022)Wallbridge, Bell, and
  Lai}]{wallbridge2022investigating}
Sarenne Wallbridge, Peter Bell, and Catherine Lai. 2022.
\newblock \href
  {https://www.isca-archive.org/interspeech_2022/wallbridge22_interspeech.pdf}
  {Investigating perception of spoken dialogue acceptability through
  surprisal}.
\newblock In \emph{Interspeech 2022: The 23rd Annual Conference of the
  International Speech Communication Association}, pages 4506--4510.
  International Speech Communication Association.

\bibitem[{Wilcox et~al.(2020)Wilcox, Gauthier, Hu, Qian, and
  Levy}]{wilcox-etal:2020-on-the-predictive-power}
Ethan~Gotlieb Wilcox, Jon Gauthier, Jennifer Hu, Peng Qian, and Roger~P. Levy.
  2020.
\newblock \href
  {https://cognitivesciencesociety.org/cogsci20/papers/0375/0375.pdf} {On the
  predictive power of neural language models for human real-time comprehension
  behavior}.
\newblock In \emph{Proceedings of the 42nd Annual Meeting of the Cognitive
  Science Society}, page 1707–1713.

\bibitem[{Wilcox et~al.(2023)Wilcox, Pimentel, Meister, Cotterell, and
  Levy}]{wilcox-2023-testing}
Ethan~Gotlieb Wilcox, Tiago Pimentel, Clara Meister, Ryan Cotterell, and
  Roger~P. Levy. 2023.
\newblock \href {https://doi.org/10.1162/tacl_a_00612} {{Testing the
  Predictions of Surprisal Theory in 11 Languages}}.
\newblock \emph{Transactions of the Association for Computational Linguistics},
  11:1451--1470.

\bibitem[{Xu et~al.(2023)Xu, Chon, Liu, and Futrell}]{xu-etal-2023-linearity}
Weijie Xu, Jason Chon, Tianran Liu, and Richard Futrell. 2023.
\newblock \href {https://doi.org/10.18653/v1/2023.findings-emnlp.1052} {The
  linearity of the effect of surprisal on reading times across languages}.
\newblock In \emph{Findings of the Association for Computational Linguistics:
  EMNLP 2023}, pages 15711--15721, Singapore. Association for Computational
  Linguistics.

\bibitem[{Xu and Reitter(2016)}]{xu-reitter-2016-entropy}
Yang Xu and David Reitter. 2016.
\newblock \href {https://doi.org/10.18653/v1/P16-1051} {Entropy converges
  between dialogue participants: Explanations from an information-theoretic
  perspective}.
\newblock In \emph{Proceedings of the 54th Annual Meeting of the Association
  for Computational Linguistics (Volume 1: Long Papers)}, pages 537--546,
  Berlin, Germany. Association for Computational Linguistics.

\bibitem[{Xu and Reitter(2018)}]{xu-reitter-2018-information}
Yang Xu and David Reitter. 2018.
\newblock \href
  {https://doi.org/https://doi.org/10.1016/j.cognition.2017.09.018}
  {Information density converges in dialogue: Towards an information-theoretic
  model}.
\newblock \emph{Cognition}, 170:147--163.

\bibitem[{Xu et~al.(2024)Xu, Wang, An, Liu, and Li}]{xu-etal-2024-detecting}
Yang Xu, Yu~Wang, Hao An, Zhichen Liu, and Yongyuan Li. 2024.
\newblock \href {https://doi.org/10.18653/v1/2024.emnlp-main.564} {Detecting
  subtle differences between human and model languages using spectrum of
  relative likelihood}.
\newblock In \emph{Proceedings of the 2024 Conference on Empirical Methods in
  Natural Language Processing}, pages 10108--10121, Miami, Florida, USA.
  Association for Computational Linguistics.

\bibitem[{Yang et~al.(2023)Yang, Yuan, Xu, Zhan, Bai, and Chen}]{yang2023face}
Zuhao Yang, Yingfang Yuan, Yang Xu, Shuo Zhan, Huajun Bai, and Kefan Chen.
  2023.
\newblock \href
  {https://proceedings.neurips.cc/paper_files/paper/2023/file/37094fdc81632915a5738293cf9b7ad4-Paper-Conference.pdf}
  {Face: {E}valuating natural language generation with {F}ourier analysis of
  cross-entropy}.
\newblock In \emph{Proceedings of the 37th International Conference on Neural
  Information Processing Systems}, pages 17038--17056.

\bibitem[{Yee et~al.(2024)Yee, Giulianelli, and
  Sinclair}]{yee-etal-2024-efficiency}
Jun~Sen Yee, Mario Giulianelli, and Arabella~J. Sinclair. 2024.
\newblock \href {https://aclanthology.org/2024.lrec-main.494/} {Efficiency and
  effectiveness in task-oriented dialogue: On construction repetition,
  information rate, and task success}.
\newblock In \emph{Proceedings of the 2024 Joint International Conference on
  Computational Linguistics, Language Resources and Evaluation (LREC-COLING
  2024)}, pages 5562--5577, Torino, Italia. ELRA and ICCL.

\bibitem[{Zipf(1949)}]{zipf1949human}
George~Kingsley Zipf. 1949.
\newblock \emph{Human Behavior and the Principle of Least Effort}.
\newblock Addison-Wesley Press, Oxford, England.

\end{thebibliography}
\bibliographystyle{acl_natbib}

\clearpage
\appendix

\section{Reproducibility}
We conduct sentence segmentation on the Spanish RST Discourse Treebank with the \texttt{mediacloud}\footnote{\url{https://github.com/mediacloud/sentence-splitter}.} text-to-sentence splitter and on the German Potsdam Commentary Corpus with \texttt{wtpsplit}\footnote{\url{https://github.com/segment-any-text/wtpsplit}.} \citep{minixhofer-etal-2023-wheres, frohmann-etal-2024-segment}. We recover text boundaries for the English RST Discourse Treebank using the corresponding texts in the Penn Treebank \citep{marcus1999treebank}.\looseness-1

To estimate surprisal for the English RST Discourse Treebank and the Spanish RST Discourse Treebank, we follow \citet{tsipidi-etal-2024-surprise} and use an RTX 4090 GPU with VRAM 24GB and additional RAM of 64GB for approximately 6 hours; for the Brazilian Portuguese CST-News corpus, we use the same setup for 34 minutes. For the German Potsdam Commentary Corpus, the Dutch corpus, and the RST Basque Treebank, we run inference on an RTX A6000 GPU for circa 3 hours.\looseness-1

Our harmonic regression experiments are implemented with the Statsmodels package\footnote{\url{https://www.statsmodels.org}.} \citep{seabold-proc-scipy-2010}. They amount to approximately 37 days of compute time on CPU with 256 GB RAM (without a GPU).\looseness-1

\section{Datasets \& Models}
\label{sec:dataset-stats}
For English, we use the RST Discourse Treebank \citep{carlson-etal-2001-building, carlson-marcu-2001-discourse}, which consists of 347 documents from the Wall Street Journal. For Spanish, we use the Spanish RST Discourse Treebank \citep{da-cunha-etal-2011-development}, containing 267 documents from a variety of domains, including astrophysics, mathematics, and law.\footnote{After removing 11 documents with missing nodes, we retain 256 documents.} For German, we turn to the Potsdam Commentary Corpus 2.0 \citep{stede-2004-potsdam, stede-neumann-2014-potsdam}, which consists of 176 documents annotated with 30 discourse relations under the RST framework. The Dutch corpus \citep{vliet-etal-2011-building, redeker-etal-2012-multi} contains 80 documents from various sources such as science news, encyclopedias, fundraising letters, and commercial advertisements, annotated with 31 relations. We exclude documents with overlapping EDUs, reducing the dataset to 62 documents. For Basque, we process data from the RST Basque TreeBank\footnote{\href{https://ixa2.si.ehu.eus/diskurtsoa/en/}{https://ixa2.si.ehu.eus/diskurtsoa/en/}}
\citep{iruskieta-etal-2013-RSTBasque}, consisting of 88 abstracts from medicine, terminology, and science articles, which are annotated with 31 relations. Finally, the CST-News corpus \citep{Cardoso-etal-2011} includes 140 Brazilian Portuguese news documents annotated with 31 relations. We remove 14 documents that have a mismatch between the raw text and the RSTs, leading to a total of 126 documents.

\Cref{tab:datstats} presents the number of documents, paragraphs, sentences, and EDUs for each language, while \Cref{tab:stats} and \Cref{tab:stats2} provide the token counts per EDU, sentence, and paragraph. These values vary considerably between languages. For example, English has a median of 11 tokens per EDU, compared to a higher median of 26 tokens in Basque.
\begin{table}[!ht]
\centering
\small
\setlength{\tabcolsep}{4pt}
\begin{tabular}{lrrrr}
\toprule
\textbf{Dataset} & \textbf{\#Docs} & \textbf{\#Pars} & \textbf{\#Sents} & \textbf{\#EDUs} \\
\midrule
English & 347 & 3511 & 7012 & 19443 \\
Spanish & 256 & 963 & 2065 & 3146 \\
German & 176 & 531 & 2097 & 3018 \\
Dutch & 62 & 371 & 1310 & 1761 \\
Basque & 88 & 198 & 1413 & 2509 \\
Brazilian Portuguese & 126 & 927 & 1815 & 4847 \\
\bottomrule
\end{tabular}
\caption{Number of documents, paragraphs, sentences, and EDUs for each dataset.}
\label{tab:datstats}
\end{table}

\begin{table*}[h] 
\small
\centering 
\begin{tabular}{l|rrr|rrr|rrr}
\toprule
\multicolumn{1}{l}{~} & \multicolumn{3}{c|}{\textbf{English}} & \multicolumn{3}{c}{\textbf{German}} & \multicolumn{3}{|c}{\textbf{Spanish}} \\ \midrule
Category & Mean & Variance & Median & Mean & Variance & Median & Mean & Variance & Median \\
\midrule
Tokens per EDU & 12.87 & 9.13 & 11 & 19.30 & 11.29 & 17 & 30.04 & 20.75 & 25 \\
Tokens per sentence & 35.67 & 18.94 & 33 & 27.78 & 17 & 25 & 45.76 & 29.18 & 40 \\
Tokens per paragraph & 71.25 & 40.76 & 65 & 109.70 & 106.02 & 91 & 98.23 & 104.05 & 62 \\
Tokens per document & 720.89 & 622.55 & 523 & 330.98 & 29.40 & 333 & 369.15 & 291.28 & 301 \\  \midrule
EDUs per sentence & 2.77 & 1.78 & 2 & 1.44 & 0.74 & 1 & 1.52 & 0.94 & 1 \\
EDUs per paragraph & 5.54 & 3.58 & 5 & 5.68 & 5.07 & 4 & 3.27 & 3.13 & 2 \\
EDUs per document & 56.03 & 51.41 & 40 & 17.15 & 3.06 & 17 & 12.29 & 9.68 & 10 \\ \midrule
Sentences per paragraph & 2 & 1.17 & 2 & 3.95 & 3.32 & 3 & 2.15 & 1.78 & 1 \\
Sentences per document & 20.21 & 17.93 & 14 & 11.91 & 2.45 & 12 & 8.07 & 5.99 & 6 \\ \midrule
Paragraphs per document & 10.12 & 7.65 & 8 & 3.02 & 1 & 3 & 3.76 & 3.26 & 2 \\
\bottomrule
\end{tabular}
\caption{Number of tokens, EDUs, sentences, and paragraphs per unit for English, German, and Spanish.} \label{tab:stats}
 \end{table*}

 \begin{table*}[h] 
\small
\centering 
\begin{tabular}{l|rrr|rrr|rrr}
\toprule
\multicolumn{1}{l}{~} & \multicolumn{3}{c|}{\textbf{Basque}} & \multicolumn{3}{c}{\textbf{Brazilian Portuguese}} & \multicolumn{3}{|c}{\textbf{Dutch}} \\ \midrule
Category & Mean & Variance & Median & Mean & Variance & Median & Mean & Variance & Median \\
\midrule
Tokens per EDU & 31.34 & 22.93 & 26 & 16.34 & 10.41 & 14 & 20.08 & 10.92 & 18 \\
Tokens per sentence & 55.65 & 34.84 & 49 & 43.65 & 24.66 & 40 & 26.99 & 14.96 & 25 \\
Tokens per paragraph & 397.15 & 407.25 & 233 & 85.46 & 73.58 & 74 & 95.31 & 77.43 & 91 \\
Tokens per document & 893.59 & 396.58 & 903 & 638.86 & 301.70 & 610 & 570.34 & 127.63 & 572 \\ \midrule
EDUs per sentence & 1.78 & 1.13 & 1 & 2.67 & 1.85 & 2 & 1.34 & 0.63 & 1 \\
EDUs per paragraph & 12.67 & 12.09 & 8 & 5.23 & 5.16 & 4 & 4.75 & 3.53 & 4 \\
EDUs per document & 28.51 & 14.37 & 27 & 39.09 & 20.01 & 35 & 28.40 & 5.76 & 28 \\ \midrule
Sentences per paragraph & 7.14 & 6.47 & 5 & 1.96 & 1.68 & 2 & 3.53 & 2.45 & 3 \\
Sentences per document & 16.06 & 7.69 & 16 & 14.64 & 7.93 & 13 & 21.13 & 4.63 & 21 \\ \midrule
Paragraphs per document & 2.25 & 2.12 & 1 & 7.48 & 3.90 & 7 & 5.98 & 1.99 & 6 \\
\bottomrule
\end{tabular}
\caption{Number of tokens, EDUs, sentences, and paragraphs per unit for Basque, Brazilian Portuguese, and Dutch.} \label{tab:stats2}
 \end{table*}

Additionally, \Cref{tab:model-stats} shows the respective model used to estimate surprisal in each language.

\begin{table*}[h]
\centering
\small
\begin{tabular}{@{}llp{6cm}@{}}
\toprule
\textbf{Language} & \textbf{Model} & \textbf{Source} \\
\midrule
English & \textsc{Nous-Yarn-Llama-2-7b-64k} \citep{peng2024yarn} & \url{https://huggingface.co/NousResearch/Yarn-Llama-2-7b-64k} \\
Spanish & \textsc{LINCE Mistral 7B Instruct} & \url{https://huggingface.co/clibrain/lince-mistral-7b-it-es} \\
German & \textsc{LAION LeoLM 7B} & \url{https://huggingface.co/LeoLM/leo-hessianai-7b} \\
Basque & \textsc{Latxa 7B}  \citep{etxaniz2024latxa} & \url{https://huggingface.co/HiTZ/latxa-7b-v1.2} \\
Dutch & \textsc{GEITje 7B Ultra} \citep{vanroy2024geitje7bultraconversational} & \url{https://huggingface.co/BramVanroy/GEITje-7B-ultra} \\
Brazilian Portuguese & \textsc{Sabi{\'a}-7B} \citep{pires2023sabia} & \url{https://huggingface.co/maritaca-ai/sabia-7b} \\
\bottomrule
\end{tabular}
\caption{Language models used to estimate surprisal for the six languages analyzed in this study.}
\label{tab:model-stats}
\end{table*}

\section{Other Approaches to Surprisal Contour Modeling}
\label{sec:appendix_other_surprisal_modeling}
Several methods have been proposed for modeling surprisal contours in discourse, each offering different advantages in analyzing structural patterns. 
Here, we discuss three common approaches and contrast them with our use of harmonic regression. In particular, we highlight how different approaches handle positional predictors and their role in shaping surprisal curves.

\paragraph{Discrete Fourier Transform.}
Other studies analyzing the surprisal contours of discourse in the frequency domain usually decompose surprisal sequences using the discrete Fourier transform (DFT).
This yields coefficients for the whole frequency band which can be useful, e.g., for comparing the frequency profiles of human and language model-generated text \citep{yang2023face, xu-etal-2024-detecting}.  
Since we focus on surprisal curves of human discourse, we opted instead for harmonic regression, which easily allows us to identify the most significant frequencies of the signal through significance testing.

\paragraph{Contour Standardization.} 
Before applying the DFT, \citet{xu-etal-2024-detecting} additionally standardize the surprisal values (centering values around the mean with unit standard deviation) to facilitate comparisons between humans and different LLMs. 
We do not apply standardization because our goal is not to compare surprisal curves across models. Moreover, we find that standardization hinders the discovery of significant frequency components.

\paragraph{Linear Modeling.}
Early studies of surprisal contours model the surprisal of a sentence as a linear function of its absolute position with the help of n-gram models 
\citep{genzel-charniak-2002-entropy, keller-2004-entropy} or, more recently, transformers \citep{giulianelli-fernandez-2021-analysing}.
To account for the effects of sentence length, some studies additionally divide by the average surprisal of all the sentences of equal length
\citep{genzel-charniak-2003-variation, xu-reitter-2016-entropy, xu-reitter-2018-information}, an adjustment deemed less crucial when dealing with word or subword tokens \citep{verma-etal-2023-revisiting}. 
In studies where contextual structures are considered, predictors are either the absolute position of the unit within its containing structure \cite{giulianelli-etal-2021-information,maes2022shared}, or the unit's relative position \citep{tsipidi-etal-2024-surprise}.
In the present work, we apply harmonic regression---a variant of linear regression---because it allows us to model surprisal contours as a sum of sinusoidal components.\looseness-1

\section{Regularization}
\label{sec:reg-details}
To perform feature selection, we use $L_1$ regularization with a penalty weight\footnote{The Statsmodels package documentation refers to this penalty weight as $\alpha$, but we refer to it as $\penaltyWeight$ to prevent confusion with the significance level $\alpha$.} of $\penaltyWeight=0.01$, chosen by the lowest mean-squared error (MSE) on one cross-validation fold for each of the six corpora. The Brazilian Portuguese corpus is the exception, with the optimal weight for the cross-validation fold being $\penaltyWeight=0.1$; however, the difference in MSE was minimal (8.988 for $\penaltyWeight=0.1$ and 9.057 for  $\penaltyWeight=0.01$) and we opted to train on it with the same $\penaltyWeight$ as the other corpora for consistency.

\section{Baseline}
\label{sec:baseline-details}
We set up our baseline as a linear regression model trained on intercept, length of the base unit $\sym_t$, i.e., number of characters in the BPE token, surprisal of previous unit $\surprisal(\sym_{t-1}; \str_{<t-1})$, relative position of $\sym_t$ in the document, and boolean feature vectors that indicate whether $\sym_t$ is within windows of 1, 2, and 4 tokens distance from a structural boundary. Against this baseline, we run one-way ANOVA with additional models trained simultaneously on baseline features and each order of harmonic components, setting the significance level $\alpha=0.001$.
We choose a low threshold to ensure we retain features that are highly statistically significant.\looseness-1

\subsection{Baseline Coefficients in the Maximal Model}
\Cref{tab:coeff-bl-maximal-model} shows the coefficients ($\beta$) of the baseline features in the maximal model. Among these features, the Boolean boundary flags (Boundary $\pm$ 1) consistently have the highest coefficients.
\begin{table}[ht!]
    \small
    \centering
    \setlength{\tabcolsep}{4pt}
\begin{tabular}{l@{\hspace{0.50em}}c|l@{\hspace{0.50em}}c}
\toprule
\multicolumn{2}{c}{\textbf{English}} & \multicolumn{2}{c}{\textbf{Spanish}} \\
Feature & $\beta$ & Feature & $\beta$  \\ \cmidrule(lr){1-2} \cmidrule(lr){3-4}
Boundary $\pm$ 1 & 0.817$_{\texttt{10}}$ & Boundary $\pm$ 1 & 3.976$_{\texttt{10}}$ \\
Boundary $\pm$ 2 & 0.300$_{\texttt{10}}$ & Boundary $\pm$ 2 & 0.625$_{\texttt{10}}$ \\
Boundary $\pm$ 4 & - & Boundary $\pm$ 4 & 0.201$_{\texttt{10}}$ \\
Previous Surprisal & - & Previous Surprisal & 0.026$_{\texttt{10}}$ \\
Relative Position & -0.708$_{\texttt{10}}$ & Relative Position & -1.050$_{\texttt{10}}$ \\
Token Length & 0.390$_{\texttt{10}}$ & Token Length & 0.283$_{\texttt{10}}$ \\
\midrule
\midrule
\multicolumn{2}{c}{\textbf{German}} & \multicolumn{2}{c}{\textbf{Dutch}} \\
Feature & $\beta$  & Feature & $\beta$  \\ \cmidrule(lr){1-2} \cmidrule(lr){3-4}
Boundary $\pm$ 1 & 1.903$_{\texttt{10}}$ & Boundary $\pm$ 1 & 2.883$_{\texttt{10}}$ \\
Boundary $\pm$ 2 & - & Boundary $\pm$ 2 & - \\
Boundary $\pm$ 4 & - & Boundary $\pm$ 4 & - \\
Previous Surprisal & - & Previous Surprisal & 0.028$_{\texttt{10}}$ \\
Relative Position & -0.350$_{\texttt{10}}$ & Relative Position & -0.287$_{\texttt{10}}$ \\
Token Length & 0.176$_{\texttt{10}}$ & Token Length & 0.109$_{\texttt{10}}$ \\
\midrule
\midrule
\multicolumn{2}{c}{\textbf{Basque}} & \multicolumn{2}{c}{\textbf{Brazilian Portuguese}} \\
\midrule
Feature & $\beta$  & Feature & $\beta$  \\ \cmidrule(lr){1-2} \cmidrule(lr){3-4}
Boundary $\pm$ 1 & 3.699$_{\texttt{10}}$ & Boundary $\pm$ 1 & 0.982$_{\texttt{10}}$ \\
Boundary $\pm$ 2 & -0.639$_{\texttt{10}}$ & Boundary $\pm$ 2 & 0.468$_{\texttt{10}}$ \\
Boundary $\pm$ 4 & -0.152$_{\texttt{10}}$ & Boundary $\pm$ 4 & - \\
Previous Surprisal & - & Previous Surprisal & -0.015$_{\texttt{10}}$ \\
Relative Position & - & Relative Position & - \\
Token Length & 0.230$_{\texttt{10}}$ & Token Length & 0.298$_{\texttt{10}}$ \\
\bottomrule
\end{tabular}
\caption{Mean coefficients $\beta$ of the baseline predictors in the maximal model (All). Subscripts denote the number of cross-validation folds (out of a total of ten folds) where the sinusoids are significant in the ANOVA. Missing coefficient values (-) indicate features that do not persist through feature selection in all ten cross-validation folds.}
\label{tab:coeff-bl-maximal-model}
\vspace{-1em}
\end{table}

\subsection{Baseline Coefficients in the Baseline Model}
\label{sec:baseline_results}

In \Cref{tab:baseline-features}, we compare the coefficients ($\beta$) of the baseline features in the baseline model. Across languages, the Boolean feature vectors of size 1 (Boundary $\pm$ 1) around the token consistently have the largest coefficient magnitude.

\begin{table}[ht!]
\small
\setlength{\tabcolsep}{4pt}
\begin{tabular}{l@{\hspace{0.50em}}c|l@{\hspace{0.50em}}c}
\toprule
\multicolumn{2}{c}{\textbf{English}} & \multicolumn{2}{c}{\textbf{Spanish}} \\\cmidrule(lr){1-2} \cmidrule(lr){3-4}
Feature & $\beta$ & Feature & $\beta$ \\
\midrule
Boundary $\pm$ 1 & 0.899$_{\texttt{9}}$ & Boundary $\pm$ 1 & 2.713$_{\texttt{10}}$ \\
Boundary $\pm$ 2 & - &  Boundary $\pm$ 2 & 0.378$_{\texttt{10}}$ \\
Boundary $\pm$ 4 & - &  Boundary $\pm$ 4 & -0.254$_{\texttt{10}}$ \\
Previous Surprisal & - & Previous Surprisal & 0.047$_{\texttt{10}}$ \\
Relative Position & -0.415$_{\texttt{9}}$ & Relative Position & 1.356$_{\texttt{10}}$ \\
 Token length & 0.401$_{\texttt{8}}$  & Token length & 0.315$_{\texttt{10}}$ \\
\midrule
\midrule
\multicolumn{2}{c}{\textbf{German}} & \multicolumn{2}{c}{\textbf{Dutch}} \\\cmidrule(lr){1-2} \cmidrule(lr){3-4}
Feature & $\beta$ & Feature & $\beta$ \\
\midrule
Boundary $\pm$ 1 & 1.680$_{\texttt{9}}$ & Boundary $\pm$ 1 & 2.082$_{\texttt{10}}$ \\
Boundary $\pm$ 2 & - & Boundary $\pm$ 2 & -  \\
Boundary $\pm$ 4 & -0.189$_{\texttt{9}}$ & Boundary $\pm$ 4 & - \\
Previous Surprisal & 0.079$_{\texttt{8}}$ & Previous Surprisal & 0.083$_{\texttt{9}}$ \\
Relative Position & -0.980$_{\texttt{9}}$ & Relative Position & -0.688$_{\texttt{10}}$ \\
Token length & 0.248$_{\texttt{8}}$ & Token length & 0.140$_{\texttt{8}}$ \\
\midrule
\midrule
\multicolumn{2}{c}{\textbf{Basque}} & \multicolumn{2}{c}{\textbf{Brazilian Portuguese}} \\\cmidrule(lr){1-2} \cmidrule(lr){3-4}
Feature & $\beta$ & Feature & $\beta$ \\
\midrule
Boundary $\pm$ 1 & 2.677$_{\texttt{10}}$ & Boundary $\pm$ 1 & 1.003$_{\texttt{10}}$ \\
Boundary $\pm$ 2 & -0.686$_{\texttt{10}}$ & Boundary $\pm$ 2 & 0.196$_{\texttt{10}}$ \\
Boundary $\pm$ 4 & -0.190$_{\texttt{7}}$ & Boundary $\pm$ 4 & - \\
Previous Surprisal & 0.054$_{\texttt{9}}$ & Previous Surprisal & - \\
Relative Position & -0.365$_{\texttt{10}}$ & Relative Position & -0.301$_{\texttt{10}}$ \\
Token length & 0.267$_{\texttt{10}}$ &  Token length & 0.318$_{\texttt{10}}$  \\
\bottomrule
\end{tabular}
\caption{Mean coefficients ($\beta$) of the baseline features in the baseline model. Subscripts denote the number of cross-validation folds (out of a total of ten folds) where the sinusoids are significant in the ANOVA. Missing coefficient values (-) indicate features that do not persist through feature selection in all ten cross-validation folds.}
\label{tab:baseline-features}
\end{table}

\section{Significance Testing}
\label{sec:significance-baseline}
In \Cref{tab:sig_results}, we report the significance of the MSE reduction over the baseline for each scaling method and language using a paired, one-sided $t$-test. For each language, we have ten validation folds, hence ten paired ($n = 10$) observations between the baseline and each scaling method. We test whether each scaling method reduces the MSE with a one-sided paired $t$-test\footnote{\url{https://docs.scipy.org/doc/scipy/reference/generated/scipy.stats.ttest_rel.html}.}
($H_{0}: \mu_{\text{scaled}}\!\ge\!\mu_{\text{base}}$,
$H_{1}: \mu_{\text{scaled}}\!<\!\mu_{\text{base}}$). To account for the multiple comparisons, we we the Holm–Bonferroni correction\footnote{\url{https://www.statsmodels.org/dev/generated/statsmodels.stats.multitest.multipletests.html}.} \citep{holm1979simple} and report both original and corrected $\significance$-values in \Cref{tab:sig_results}.

\begin{table*}[t]
\small
\centering
\begin{tabular}{llccc}
\toprule
\textbf{Language} & \textbf{Setting} & \textbf{$\Delta$MSE $\downarrow$} & $\significance_{raw}$ &$\significance_{Holm}$\\
\midrule
English & All & 0.54* & 5.61e-12 & 1.63e-10 \\
English & Document-scaled & 0.00 & 0.727 & 1 \\
English & EDU-scaled & 0.45* & 2.77e-11 & 7.47e-10 \\
English & Paragraph-scaled & 0.19* & 8.04e-10 & 1.85e-08 \\
English & Sentence-scaled & 0.36* & 1.61e-11 & 4.5e-10 \\
\midrule
Spanish & All & 1.54* & 1.01e-07 & 1.72e-06 \\
Spanish & Document-scaled & 1.11* & 1.92e-06 & 2.76e-05 \\
Spanish & EDU-scaled & 0.80* & 2.2e-09 & 4.83e-08 \\
Spanish & Paragraph-scaled & 0.23* & 1.1e-05 & 0.000131 \\
Spanish & Sentence-scaled & 0.46* & 1.17e-08 & 2.35e-07 \\
\midrule
German & All & 1.05* & 4.05e-10 & 9.71e-09 \\
German & Document-scaled & 0.14* & 1.81e-05 & 0.0002 \\
German & EDU-scaled & 1.11* & 3.36e-12 & 1.01e-10 \\
German & Paragraph-scaled & 0.19* & 1.84e-06 & 2.76e-05 \\
German & Sentence-scaled & 0.86* & 7.35e-11 & 1.84e-09 \\
\midrule
Dutch & All & 0.10 & 0.0271 & 0.217 \\
Dutch & Document-scaled & -0.28 & 1 & 1 \\
Dutch & EDU-scaled & 0.59* & 1.91e-08 & 3.63e-07 \\
Dutch & Paragraph-scaled & 0.01 & 0.319 & 1 \\
Dutch & Sentence-scaled & 0.40* & 1.99e-06 & 2.76e-05 \\
\midrule
Basque & All & -0.09 & 0.975 & 1 \\
Basque & Document-scaled & -0.17 & 1 & 1 \\
Basque & EDU-scaled & 0.33* & 2.31e-09 & 4.85e-08 \\
Basque & Paragraph-scaled & -0.24 & 1 & 1 \\
Basque & Sentence-scaled & 0.15* & 5.62e-07 & 8.99e-06 \\
\midrule
Brazilian Portuguese & All & 0.21* & 9.79e-05 & 0.000979 \\
Brazilian Portuguese & Document-scaled & -0.17 & 1 & 1 \\
Brazilian Portuguese & EDU-scaled & 0.56* & 3.25e-11 & 8.44e-10 \\
Brazilian Portuguese & Paragraph-scaled & 0.08 & 0.00641 & 0.0577 \\
Brazilian Portuguese & Sentence-scaled & 0.31* & 9.47e-08 & 1.7e-06 \\
\bottomrule
\end{tabular}
\caption{One-sided paired $t$-test ("greater") comparing each scaling method to the baseline. Holm-adjusted $\significance$ values control the family-wise error rate. Asterisks (*) indicate $\significance < .001$.}
\label{tab:sig_results}
\end{table*}

\section{Permuted Surprisal}
\label{sec:permuted_surprisal}
We replicate the experiments from \Cref{sec:empirical-findings}, this time with randomly permuted surprisal values. As shown in \Cref{tab:mse_permuted}, MSE values are generally higher compared to the results in \Cref{tab:val_mse}. Additionally, we observe no notable difference between different scaling methods.

\begin{table*}[ht!]
\small
\centering
\begin{tabular}{lcccccc}
\toprule
 & \textbf{English} & \textbf{Spanish} & \textbf{German} & \textbf{Dutch} & \textbf{Basque} & \textbf{Brazilian Portuguese}\\
\midrule
Document-scaled & $10.81\pm.45$ & $15.60\pm.50$ & $12.91\pm.28$ & $9.88\pm.85$ & $9.36\pm.55$ & $10.02\pm.84$ \\
EDU-scaled & $10.80\pm.45$ & $15.51\pm.48$ & $12.86\pm.27$ & $9.78\pm.85$ & $9.32\pm.56$ & $9.98\pm.84$ \\
Sentence-scaled & $10.81\pm.45$ & $15.51\pm.48$ & $12.86\pm.27$ & $9.79\pm.84$ & $9.33\pm.56$ & $9.98\pm.84$ \\
Paragraph-scaled & $10.81\pm.45$ & $15.53\pm.48$ & $12.89\pm.27$ & $9.84\pm.85$ & $9.36\pm.57$ & $10.00\pm.83$ \\
\midrule
All & $10.81\pm.45$ & $15.65\pm.49$ & $12.96\pm.27$ & $9.97\pm.84$ & $9.41\pm.56$ & $10.06\pm.84$ \\
\bottomrule
\end{tabular}
\caption{Mean and standard deviation for 10-fold validation MSEs across scaling settings and languages for permuted surprisal values. We observe no notable differences between different settings.}
    \label{tab:mse_permuted}
\end{table*}

\begin{figure*}[ht!]
    \centering
    \includegraphics[width=\textwidth]{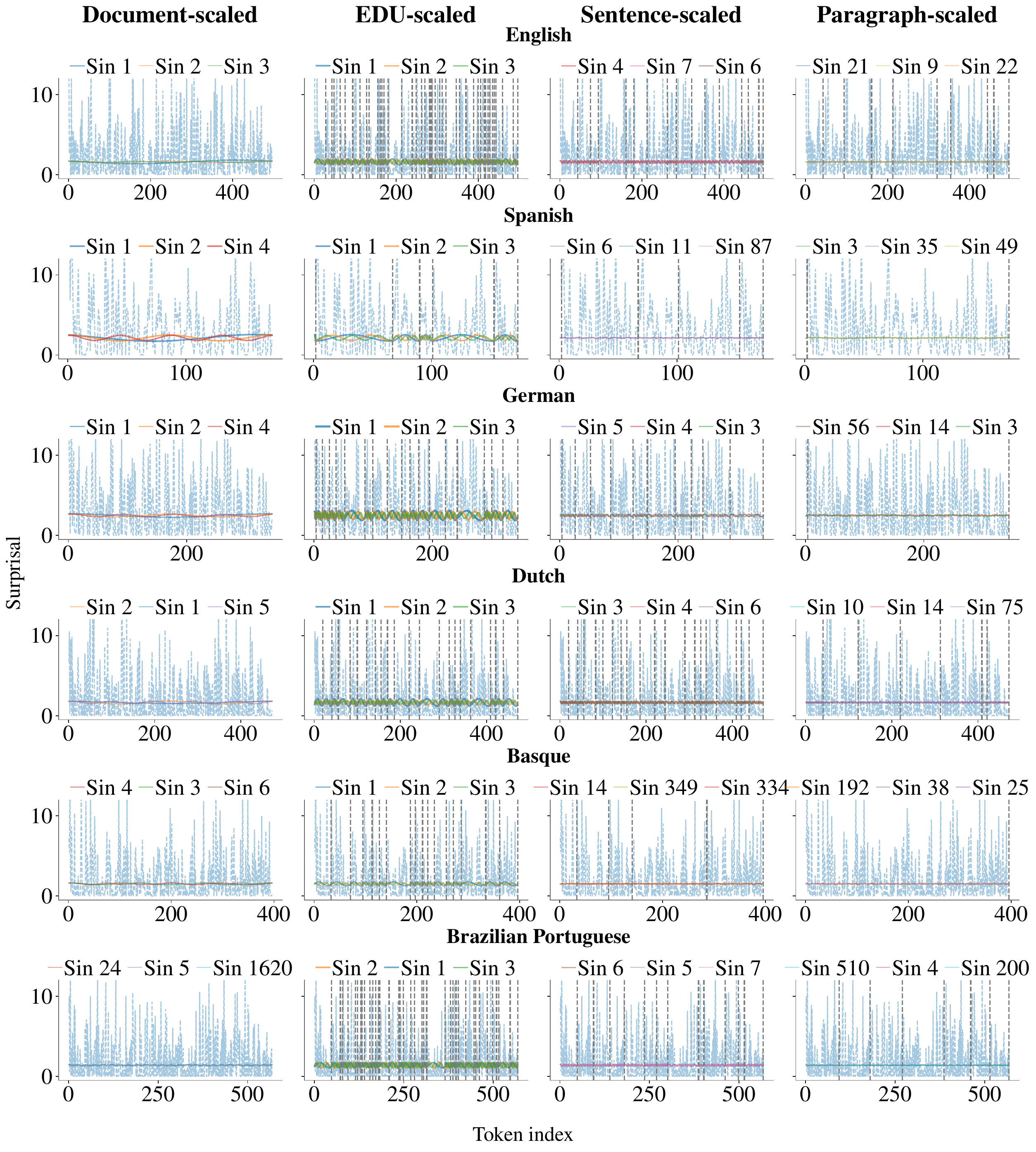}
    \caption{Top three most dominant sinusoids for English (doc \texttt{wsj\_1111}), Spanish (doc \texttt{ec00002}), German (doc \texttt{maz-11280}), Dutch (doc \texttt{FL13\_Unicef}), Basque (doc \texttt{TERM29-GS}), Brazilian Portuguese (doc \texttt{D2\_C38\_Estadao}). Amplitudes signify the contribution to the overall variation, with higher amplitudes indicating a larger effect.
    }
    \label{fig:predictive-harmonics-vis-app-0}
\end{figure*}

\begin{figure*}[ht!]
    \centering
    \includegraphics[width=\textwidth]{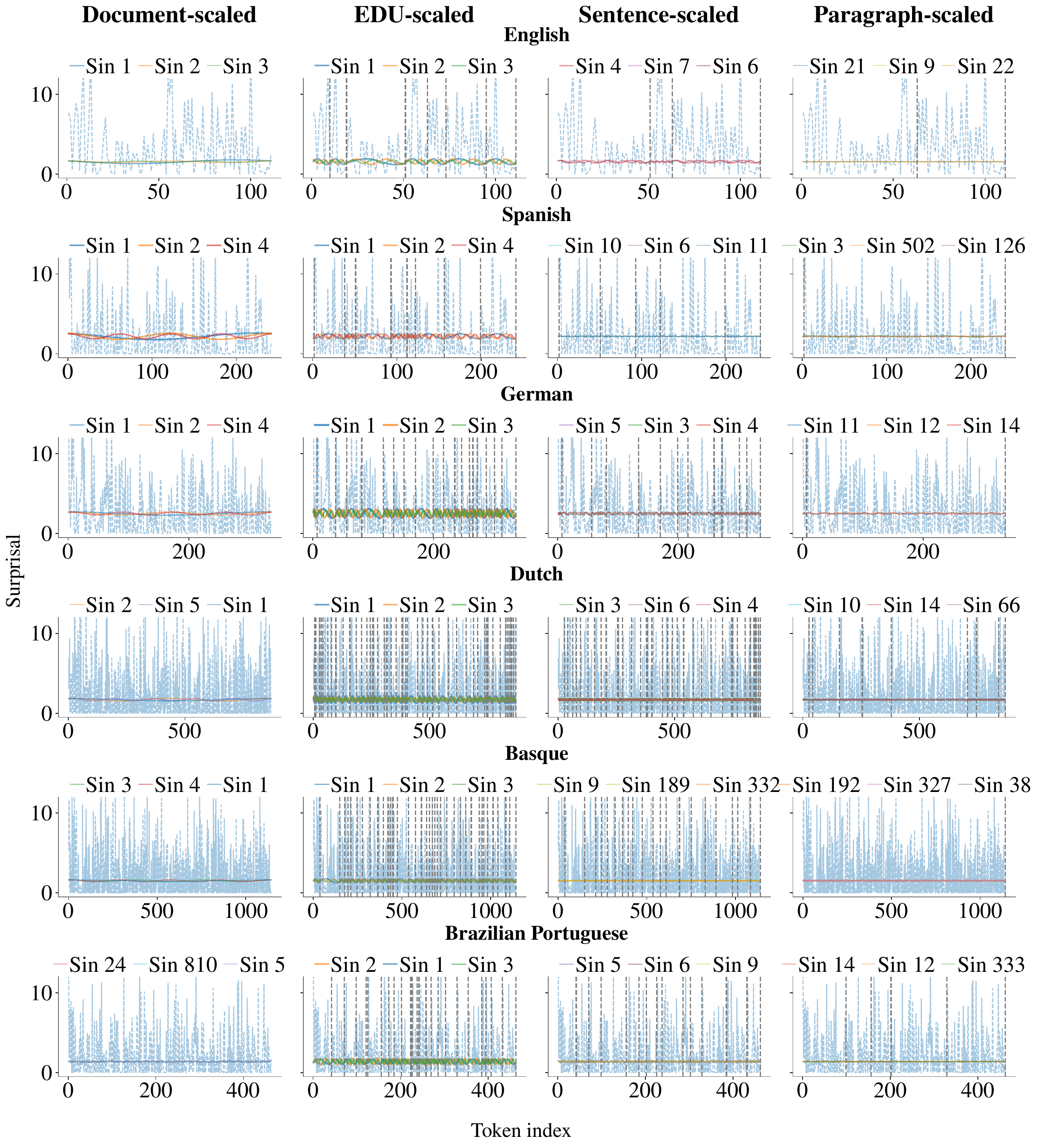}
    \caption{Top three most dominant sinusoids for English (doc \texttt{wsj\_0605}), Spanish (doc \texttt{ec00007}), German (doc \texttt{maz-11507}), Dutch (doc \texttt{AD02\_Atkins}), Basque (doc \texttt{GMB0201-GS}), Brazilian Portuguese (doc \texttt{D3\_C11\_OGlobo}). Amplitudes signify the contribution to the overall variation, with higher amplitudes indicating a larger effect.
    }
    \label{fig:predictive-harmonics-vis-app-1}
\end{figure*}

\section{Visualizations}

Here, we present additional visualizations showing the contribution of individual sinusoids in the maximal model and how well our models recover the observed surprisal curves under different scaling methods. 

\subsection{Sinusoid Visualizations}
\label{sec:sin-vis}

In \Cref{fig:predictive-harmonics-vis-app-0} and \Cref{fig:predictive-harmonics-vis-app-1}, we present visualizations of surprisal contours, unit boundaries, and the three most dominant sinusoids for individual documents in all languages. For sinusoids, higher amplitudes correspond to a greater effect on the shape of the surprisal contour. Note that the contribution of individual sinusoids is relatively small. For combined predictions across different settings, see \Cref{sec:prediction-vis}.

\subsection{Prediction Visualizations}
\label{sec:prediction-vis}
\begin{figure*}[htp!]
\centering
    \includegraphics[width=\textwidth]{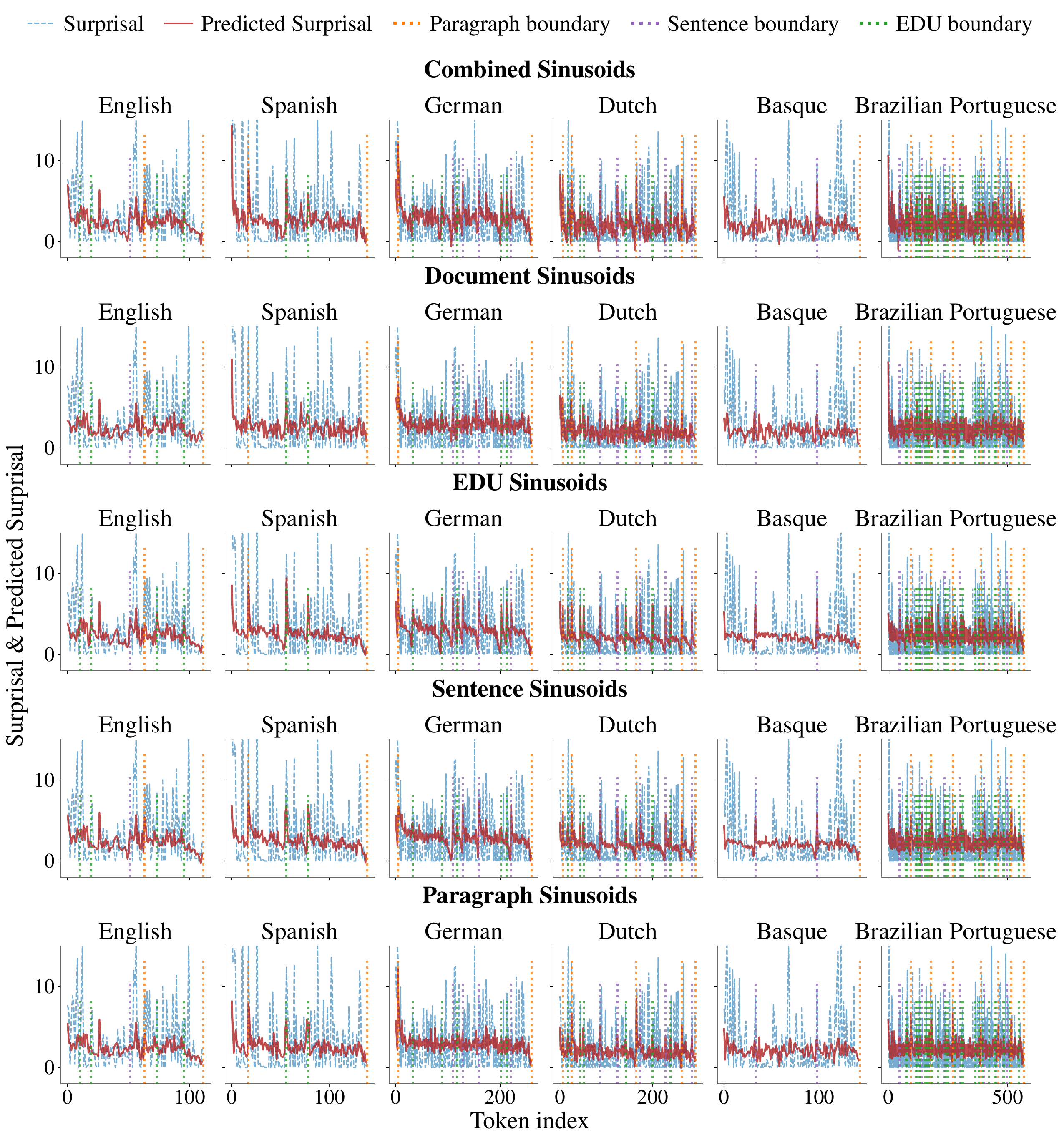}
    \caption{Predicted curves for all languages and all time-scaled sinusoids. Languages correspond to the following documents: English (\texttt{wsj\_0605}), Spanish (\texttt{as00007}), German (maz-1818), Dutch (\texttt{AD14\_CarpeDiem}), Basque (GMB0002-GS), Brazilian Portuguese (\texttt{D2\_C38\_Estadao}).
    }
    \label{fig:predicted-curves-all-settings}
\end{figure*}

Similar to the predicted curves based on EDU-scaled sinusoids in \Cref{fig:predicted-curves-edu}, we visualize the predicted curves for each scaling method in \Cref{fig:predicted-curves-all-settings}.

\section{Sinusoid Amplitudes}
\label{sec:sinusoid-amp}
In \Cref{tab:excluded-sinusoids}, we report the total number of sinusoids and the number that remain in all folds after regularization. Sinusoids that persist across all folds exhibit higher mean amplitudes ($\amp_k$) than those excluded in some folds. Additionally, in \Cref{tab:amplitudes-app}, we show the twenty-five sinusoids with the highest amplitudes that persist in all folds after $L_1$ regularization. Notably, among all settings, only the EDU-scaled sinusoids are significant across all folds in the ANOVA.

\begin{table*}
\centering
\small
\begin{tabular}{llcccc}
\toprule
  \textbf{Language}             & \textbf{Setting}          & \textbf{\# Sinusoids} &  \textbf{\# All folds} &   \textbf{Mean $\amp_k$ (All folds)} & \textbf{Mean $\amp_k$ (Excluded)} \\
\midrule
English              & Document-scaled  &        474 &         1 &         0.235 &          0.038 \\
English              & EDU-scaled       &         60 &         6 &         0.222 &          0.058 \\
English              & Sentence-scaled  &         63 &        11 &         0.099 &          0.059 \\
English              & Paragraph-scaled &        112 &         1 &         0.037 &          0.031 \\
\midrule
Spanish              & Document-scaled  &       1470 &        80 &         0.149 &          0.042 \\
Spanish              & EDU-scaled       &        116 &        13 &         0.195 &          0.050 \\
Spanish              & Sentence-scaled  &        144 &         0 &         - &          0.041 \\
Spanish              & Paragraph-scaled &        495 &         5 &         0.050 &          0.040 \\
\midrule
German               & Document-scaled  &        457 &        13 &         0.087 &          0.048 \\
German               & EDU-scaled       &         59 &        12 &         0.230 &          0.072 \\
German               & Sentence-scaled  &         98 &         5 &         0.064 &          0.061 \\
German               & Paragraph-scaled &        344 &         6 &         0.066 &          0.046 \\
\midrule
Dutch                & Document-scaled  &        817 &        27 &         0.074 &          0.044 \\
Dutch                & EDU-scaled       &         64 &        11 &         0.180 &          0.080 \\
Dutch                & Sentence-scaled  &        106 &         2 &         0.170 &          0.056 \\
Dutch                & Paragraph-scaled &        390 &        20 &         0.065 &          0.047 \\
\midrule
Basque               & Document-scaled  &       1263 &        15 &         0.062 &          0.038 \\
Basque               & EDU-scaled       &        135 &         6 &         0.132 &          0.060 \\
Basque               & Sentence-scaled  &        248 &         2 &         0.048 &          0.040 \\
Basque               & Paragraph-scaled &       1233 &         3 &         0.054 &          0.037 \\
\midrule
Brazilian Portuguese & Document-scaled  &       1225 &        13 &         0.054 &          0.040 \\
Brazilian Portuguese & EDU-scaled       &         67 &        10 &         0.188 &          0.087 \\
Brazilian Portuguese & Sentence-scaled  &        278 &         6 &         0.073 &          0.042 \\
Brazilian Portuguese & Paragraph-scaled &        635 &         2 &         0.043 &          0.038 \\
\bottomrule
\end{tabular}
\caption{Total number of sinusoids and number of sinusoids that remain in all folds after $L_1$ regularization. Sinusoids that remain in all folds exhibit higher mean amplitudes ($\amp_k$) than those excluded in some folds.}
\label{tab:excluded-sinusoids}
\end{table*}

\section{Surprisal at Boundaries}
\label{sec:boundary-surprisal-app}
We report the mean and standard deviation of token surprisal immediately before and after paragraph, sentence, and EDU boundaries. As shown in \Cref{tab:surprisal-boundaries-app}, surprisal tends to be lower before and higher after each type of boundary relative to tokens located farther from any boundary. Here, a window size of 1 means we include only the single token immediately before or after the boundary. Non-boundary tokens exclude any tokens within one position on either side of a boundary.

\begin{table*}[ht!]
\small
\centering
\begin{tabular}{l|ccccc}
\toprule
\textbf{Language} & \textbf{Window size} & \textbf{Paragraph Boundary} & \textbf{Sentence Boundar}y & \textbf{EDU Boundary} & \textbf{Non-boundary} \\
\midrule
\multicolumn{6}{c}{\textbf{Before Boundaries}} \\
\midrule
English & 1 & 1.03 $\pm$ 1.44 & 1.08 $\pm$ 1.34 & 1.79 $\pm$ 2.33 & 2.54 $\pm$ 3.24 \\
German & 1 & 1.47 $\pm$ 2.69 & 1.07 $\pm$ 1.77 & 1.30 $\pm$ 1.87 & 2.88 $\pm$ 3.53 \\
Spanish & 1 & 1.24 $\pm$ 2.10 & 1.35 $\pm$ 1.74 & 1.48 $\pm$ 1.99 & 2.44 $\pm$ 3.52 \\
Basque & 1 & 1.27 $\pm$ 2.08 & 1.32 $\pm$ 1.60 & 1.43 $\pm$ 1.68 & 2.00 $\pm$ 3.00 \\
Dutch & 1 & 1.60 $\pm$ 2.32 & 1.63 $\pm$ 2.01 & 1.53 $\pm$ 2.02 & 1.88 $\pm$ 2.99 \\
Br. Port. & 1 & 1.27 $\pm$ 1.64 & 1.37 $\pm$ 1.52 & 1.34 $\pm$ 1.69 & 2.21 $\pm$ 3.14 \\
\midrule
English & 2 & 1.26 $\pm$ 1.94 & 1.27 $\pm$ 1.88 & 1.86 $\pm$ 2.57 & 2.47 $\pm$ 3.19 \\
German & 2 & 1.88 $\pm$ 3.18 & 1.02 $\pm$ 2.05 & 1.13 $\pm$ 2.09 & 2.89 $\pm$ 3.51 \\
Spanish & 2 & 1.70 $\pm$ 4.07 & 1.31 $\pm$ 3.03 & 1.35 $\pm$ 2.90 & 2.41 $\pm$ 3.47 \\
Basque & 2 & 0.99 $\pm$ 1.78 & 0.92 $\pm$ 1.52 & 1.02 $\pm$ 1.64 & 2.05 $\pm$ 3.04 \\
Dutch & 2 & 1.24 $\pm$ 2.51 & 1.12 $\pm$ 2.11 & 1.10 $\pm$ 2.09 & 1.87 $\pm$ 2.97 \\
Brazilian Portuguese & 2 & 1.04 $\pm$ 1.88 & 1.02 $\pm$ 1.68 & 1.19 $\pm$ 2.01 & 2.17 $\pm$ 3.09 \\
\midrule
\multicolumn{6}{c}{\textbf{After Boundaries}} \\
\midrule
English & 1 & 6.34 $\pm$ 4.12 & 6.05 $\pm$ 3.80 & 4.67 $\pm$ 3.78 & 2.54 $\pm$ 3.24 \\
German & 1 & 7.75 $\pm$ 5.31 & 7.06 $\pm$ 3.73 & 6.39 $\pm$ 3.79 & 2.88 $\pm$ 3.53 \\
Spanish & 1 & 12.39 $\pm$ 12.75 & 9.06 $\pm$ 10.09 & 7.88 $\pm$ 8.77 & 2.44 $\pm$ 3.52 \\
Basque & 1 & 5.41 $\pm$ 3.69 & 5.89 $\pm$ 3.11 & 5.68 $\pm$ 3.24 & 2.00 $\pm$ 3.00 \\
Dutch & 1 & 7.80 $\pm$ 4.77 & 6.52 $\pm$ 3.84 & 5.76 $\pm$ 3.89 & 1.88 $\pm$ 2.99 \\
Br. Port. & 1 & 5.84 $\pm$ 3.86 & 5.64 $\pm$ 3.54 & 4.80 $\pm$ 3.45 & 2.21 $\pm$ 3.14 \\
\midrule
English & 2 & 5.30 $\pm$ 4.27 & 5.00 $\pm$ 4.00 & 4.29 $\pm$ 3.78 & 2.47 $\pm$ 3.19 \\
German & 2 & 7.26 $\pm$ 4.84 & 6.04 $\pm$ 3.97 & 5.61 $\pm$ 3.90 & 2.89 $\pm$ 3.51 \\
Spanish & 2 & 9.79 $\pm$ 10.58 & 7.24 $\pm$ 8.24 & 6.39 $\pm$ 7.21 & 2.41 $\pm$ 3.47 \\
Basque & 2 & 4.64 $\pm$ 3.37 & 4.09 $\pm$ 3.40 & 3.84 $\pm$ 3.49 & 2.05 $\pm$ 3.04 \\
Dutch & 2 & 6.01 $\pm$ 4.94 & 4.88 $\pm$ 4.08 & 4.54 $\pm$ 3.95 & 1.87 $\pm$ 2.97 \\
Brazilian Portuguese & 2 & 5.46 $\pm$ 3.95 & 5.02 $\pm$ 3.74 & 4.39 $\pm$ 3.60 & 2.17 $\pm$ 3.09 \\
\bottomrule
\end{tabular}
     \caption{Mean and standard deviation of token surprisal before and after paragraph, sentence, and EDU boundaries. Surprisal is lower before and higher after all boundary types compared to tokens distant from any boundary.\looseness-1}
     \label{tab:surprisal-boundaries-app}
\end{table*}

\begin{table*}[ht!]
    \centering
    \tiny
\begin{tabular}{c@{\hspace{0.50em}}cc@{\hspace{0.50em}}cc@{\hspace{0.50em}}cc@{\hspace{0.50em}}c|c@{\hspace{0.50em}}cc@{\hspace{0.50em}}cc@{\hspace{0.50em}}cc@{\hspace{0.50em}}c}
\toprule
\multicolumn{8}{c}{\textbf{English}} & \multicolumn{8}{c}{\textbf{Spanish}} \\\cmidrule(lr){1-8} \cmidrule(lr){9-16}
\multicolumn{2}{c}{Document} & \multicolumn{2}{c}{EDU} & \multicolumn{2}{c}{Sentence} & \multicolumn{2}{c}{Paragraph} & \multicolumn{2}{c}{Document} & \multicolumn{2}{c}{EDU} & \multicolumn{2}{c}{Sentence} & \multicolumn{2}{c}{Paragraph} \\
$\component$ & $\amp_\component$ & $\component$ & $\amp_\component$ & $\component$ & $\amp_\component$ & $\component$ & $\amp_\component$ & $\component$ & $\amp_\component$ & $\component$ & $\amp_\component$ & $\component$ & $\amp_\component$ & $\component$ & $\amp_\component$ \\ \cmidrule(lr){1-2} \cmidrule(lr){3-4} \cmidrule(lr){5-6} \cmidrule(lr){7-8} \cmidrule(lr){9-10} \cmidrule(lr){11-12} \cmidrule(lr){13-14} \cmidrule(lr){15-16}
1 & 0.235$_{\texttt{10}}$ & 1 & 0.370$_{\texttt{10}}$ & 4 & 0.171$_{\texttt{10}}$ & 9 & 0.037$_{\texttt{10}}$ & 1 & 0.422$_{\texttt{10}}$ & 1 & 0.364$_{\texttt{10}}$ &  &  & 126 & 0.058$_{\texttt{7}}$ \\
 &  & 2 & 0.330$_{\texttt{10}}$ & 5 & 0.151$_{\texttt{10}}$ &  &  & 4 & 0.323$_{\texttt{10}}$ & 2 & 0.313$_{\texttt{10}}$ &  &  & 535 & 0.054$_{\texttt{10}}$ \\
 &  & 3 & 0.241$_{\texttt{10}}$ & 10 & 0.144$_{\texttt{10}}$ &  &  & 5 & 0.293$_{\texttt{10}}$ & 4 & 0.264$_{\texttt{10}}$ &  &  & 150 & 0.053$_{\texttt{8}}$ \\
 &  & 4 & 0.180$_{\texttt{10}}$ & 3 & 0.134$_{\texttt{10}}$ &  &  & 7 & 0.287$_{\texttt{10}}$ & 3 & 0.261$_{\texttt{10}}$ &  &  & 381 & 0.047$_{\texttt{7}}$ \\
 &  & 5 & 0.123$_{\texttt{10}}$ & 2 & 0.112$_{\texttt{10}}$ &  &  & 6 & 0.284$_{\texttt{10}}$ & 5 & 0.235$_{\texttt{10}}$ &  &  & 173 & 0.039$_{\texttt{10}}$ \\
 &  & 7 & 0.086$_{\texttt{10}}$ & 14 & 0.092$_{\texttt{10}}$ &  &  & 8 & 0.284$_{\texttt{10}}$ & 6 & 0.205$_{\texttt{10}}$ &  &  &  &  \\
 &  &  &  & 1 & 0.066$_{\texttt{10}}$ &  &  & 10 & 0.276$_{\texttt{10}}$ & 7 & 0.205$_{\texttt{10}}$ &  &  &  &  \\
 &  &  &  & 15 & 0.060$_{\texttt{10}}$ &  &  & 9 & 0.253$_{\texttt{10}}$ & 8 & 0.159$_{\texttt{10}}$ &  &  &  &  \\
 &  &  &  & 19 & 0.059$_{\texttt{10}}$ &  &  & 13 & 0.246$_{\texttt{10}}$ & 10 & 0.140$_{\texttt{10}}$ &  &  &  &  \\
 &  &  &  & 16 & 0.056$_{\texttt{10}}$ &  &  & 15 & 0.244$_{\texttt{10}}$ & 9 & 0.139$_{\texttt{10}}$ &  &  &  &  \\
 &  &  &  & 20 & 0.046$_{\texttt{10}}$ &  &  & 11 & 0.236$_{\texttt{10}}$ & 11 & 0.107$_{\texttt{10}}$ &  &  &  &  \\
 &  &  &  &  &  &  &  & 17 & 0.223$_{\texttt{10}}$ & 13 & 0.107$_{\texttt{10}}$ &  &  &  &  \\
 &  &  &  &  &  &  &  & 21 & 0.218$_{\texttt{9}}$ & 37 & 0.036$_{\texttt{10}}$ &  &  &  &  \\
 &  &  &  &  &  &  &  & 12 & 0.210$_{\texttt{9}}$ &  &  &  &  &  &  \\
 &  &  &  &  &  &  &  & 14 & 0.207$_{\texttt{9}}$ &  &  &  &  &  &  \\
 &  &  &  &  &  &  &  & 24 & 0.206$_{\texttt{10}}$ &  &  &  &  &  &  \\
 &  &  &  &  &  &  &  & 20 & 0.205$_{\texttt{8}}$ &  &  &  &  &  &  \\
 &  &  &  &  &  &  &  & 16 & 0.203$_{\texttt{9}}$ &  &  &  &  &  &  \\
 &  &  &  &  &  &  &  & 25 & 0.197$_{\texttt{10}}$ &  &  &  &  &  &  \\
 &  &  &  &  &  &  &  & 28 & 0.197$_{\texttt{8}}$ &  &  &  &  &  &  \\
 &  &  &  &  &  &  &  & 19 & 0.195$_{\texttt{9}}$ &  &  &  &  &  &  \\
 &  &  &  &  &  &  &  & 23 & 0.194$_{\texttt{10}}$ &  &  &  &  &  &  \\
 &  &  &  &  &  &  &  & 18 & 0.194$_{\texttt{9}}$ &  &  &  &  &  &  \\
 &  &  &  &  &  &  &  & 46 & 0.191$_{\texttt{10}}$ &  &  &  &  &  &  \\
 &  &  &  &  &  &  &  & 22 & 0.190$_{\texttt{8}}$ &  &  &  &  &  &  \\
\midrule
\midrule
\multicolumn{8}{c}{\textbf{German}} & \multicolumn{8}{c}{\textbf{Dutch}} \\\cmidrule(lr){1-8} \cmidrule(lr){9-16}
\multicolumn{2}{c}{Document} & \multicolumn{2}{c}{EDU} & \multicolumn{2}{c}{Sentence} & \multicolumn{2}{c}{Paragraph} & \multicolumn{2}{c}{Document} & \multicolumn{2}{c}{EDU} & \multicolumn{2}{c}{Sentence} & \multicolumn{2}{c}{Paragraph} \\
$\component$ & $\amp_\component$ & $\component$ & $\amp_\component$ & $\component$ & $\amp_\component$ & $\component$ & $\amp_\component$ & $\component$ & $\amp_\component$ & $\component$ & $\amp_\component$ & $\component$ & $\amp_\component$ & $\component$ & $\amp_\component$ \\ \cmidrule(lr){1-2} \cmidrule(lr){3-4} \cmidrule(lr){5-6} \cmidrule(lr){7-8} \cmidrule(lr){9-10} \cmidrule(lr){11-12} \cmidrule(lr){13-14} \cmidrule(lr){15-16}
4 & 0.165$_{\texttt{10}}$ & 1 & 0.599$_{\texttt{10}}$ & 10 & 0.101$_{\texttt{10}}$ & 11 & 0.087$_{\texttt{10}}$ & 5 & 0.153$_{\texttt{10}}$ & 1 & 0.470$_{\texttt{10}}$ & 3 & 0.198$_{\texttt{10}}$ & 10 & 0.095$_{\texttt{10}}$ \\
5 & 0.140$_{\texttt{10}}$ & 2 & 0.515$_{\texttt{10}}$ & 12 & 0.063$_{\texttt{10}}$ & 56 & 0.077$_{\texttt{10}}$ & 6 & 0.135$_{\texttt{10}}$ & 2 & 0.308$_{\texttt{10}}$ & 1 & 0.141$_{\texttt{10}}$ & 66 & 0.089$_{\texttt{10}}$ \\
6 & 0.137$_{\texttt{10}}$ & 3 & 0.380$_{\texttt{10}}$ & 13 & 0.059$_{\texttt{10}}$ & 345 & 0.066$_{\texttt{9}}$ & 7 & 0.103$_{\texttt{6}}$ & 3 & 0.247$_{\texttt{10}}$ &  &  & 8 & 0.074$_{\texttt{10}}$ \\
7 & 0.112$_{\texttt{10}}$ & 4 & 0.256$_{\texttt{10}}$ & 33 & 0.049$_{\texttt{10}}$ & 57 & 0.062$_{\texttt{0}}$ & 3 & 0.102$_{\texttt{2}}$ & 7 & 0.165$_{\texttt{10}}$ &  &  & 43 & 0.072$_{\texttt{9}}$ \\
10 & 0.091$_{\texttt{10}}$ & 5 & 0.206$_{\texttt{10}}$ & 35 & 0.049$_{\texttt{10}}$ & 110 & 0.057$_{\texttt{8}}$ & 4 & 0.082$_{\texttt{2}}$ & 5 & 0.159$_{\texttt{10}}$ &  &  & 32 & 0.071$_{\texttt{10}}$ \\
8 & 0.091$_{\texttt{10}}$ & 7 & 0.181$_{\texttt{10}}$ &  &  & 167 & 0.045$_{\texttt{10}}$ & 218 & 0.082$_{\texttt{2}}$ & 4 & 0.153$_{\texttt{10}}$ &  &  & 175 & 0.068$_{\texttt{10}}$ \\
11 & 0.085$_{\texttt{10}}$ & 8 & 0.150$_{\texttt{10}}$ &  &  &  &  & 453 & 0.080$_{\texttt{0}}$ & 11 & 0.133$_{\texttt{10}}$ &  &  & 23 & 0.068$_{\texttt{0}}$ \\
22 & 0.073$_{\texttt{10}}$ & 9 & 0.135$_{\texttt{10}}$ &  &  &  &  & 8 & 0.078$_{\texttt{0}}$ & 16 & 0.121$_{\texttt{10}}$ &  &  & 4 & 0.067$_{\texttt{10}}$ \\
285 & 0.056$_{\texttt{6}}$ & 11 & 0.118$_{\texttt{10}}$ &  &  &  &  & 13 & 0.076$_{\texttt{0}}$ & 18 & 0.084$_{\texttt{10}}$ &  &  & 7 & 0.064$_{\texttt{3}}$ \\
111 & 0.053$_{\texttt{7}}$ & 14 & 0.088$_{\texttt{10}}$ &  &  &  &  & 140 & 0.076$_{\texttt{4}}$ & 6 & 0.074$_{\texttt{10}}$ &  &  & 247 & 0.063$_{\texttt{1}}$ \\
55 & 0.047$_{\texttt{1}}$ & 16 & 0.073$_{\texttt{10}}$ &  &  &  &  & 656 & 0.076$_{\texttt{0}}$ & 47 & 0.060$_{\texttt{10}}$ &  &  & 5 & 0.063$_{\texttt{8}}$ \\
86 & 0.045$_{\texttt{8}}$ & 19 & 0.053$_{\texttt{10}}$ &  &  &  &  & 9 & 0.075$_{\texttt{0}}$ &  &  &  &  & 13 & 0.061$_{\texttt{9}}$ \\
63 & 0.039$_{\texttt{8}}$ &  &  &  &  &  &  & 227 & 0.074$_{\texttt{0}}$ &  &  &  &  & 357 & 0.059$_{\texttt{3}}$ \\
 &  &  &  &  &  &  &  & 845 & 0.067$_{\texttt{0}}$ &  &  &  &  & 47 & 0.058$_{\texttt{0}}$ \\
 &  &  &  &  &  &  &  & 365 & 0.065$_{\texttt{0}}$ &  &  &  &  & 35 & 0.057$_{\texttt{9}}$ \\
 &  &  &  &  &  &  &  & 296 & 0.064$_{\texttt{0}}$ &  &  &  &  & 202 & 0.055$_{\texttt{9}}$ \\
 &  &  &  &  &  &  &  & 199 & 0.062$_{\texttt{0}}$ &  &  &  &  & 38 & 0.054$_{\texttt{10}}$ \\
 &  &  &  &  &  &  &  & 812 & 0.060$_{\texttt{0}}$ &  &  &  &  & 305 & 0.053$_{\texttt{0}}$ \\
 &  &  &  &  &  &  &  & 15 & 0.059$_{\texttt{0}}$ &  &  &  &  & 29 & 0.051$_{\texttt{10}}$ \\
 &  &  &  &  &  &  &  & 11 & 0.059$_{\texttt{0}}$ &  &  &  &  & 290 & 0.048$_{\texttt{8}}$ \\
 &  &  &  &  &  &  &  & 540 & 0.059$_{\texttt{0}}$ &  &  &  &  &  &  \\
 &  &  &  &  &  &  &  & 245 & 0.053$_{\texttt{0}}$ &  &  &  &  &  &  \\
 &  &  &  &  &  &  &  & 130 & 0.053$_{\texttt{0}}$ &  &  &  &  &  &  \\
 &  &  &  &  &  &  &  & 132 & 0.052$_{\texttt{0}}$ &  &  &  &  &  &  \\
 &  &  &  &  &  &  &  & 235 & 0.052$_{\texttt{0}}$ &  &  &  &  &  &  \\
\midrule
\midrule
\multicolumn{8}{c}{\textbf{Basque}} & \multicolumn{8}{c}{\textbf{Brazilian Portuguese}} \\\cmidrule(lr){1-8} \cmidrule(lr){9-16}
\multicolumn{2}{c}{Document} & \multicolumn{2}{c}{EDU} & \multicolumn{2}{c}{Sentence} & \multicolumn{2}{c}{Paragraph} & \multicolumn{2}{c}{Document} & \multicolumn{2}{c}{EDU} & \multicolumn{2}{c}{Sentence} & \multicolumn{2}{c}{Paragraph} \\
$\component$ & $\amp_\component$ & $\component$ & $\amp_\component$ & $\component$ & $\amp_\component$ & $\component$ & $\amp_\component$ & $\component$ & $\amp_\component$ & $\component$ & $\amp_\component$ & $\component$ & $\amp_\component$ & $\component$ & $\amp_\component$ \\ \cmidrule(lr){1-2} \cmidrule(lr){3-4} \cmidrule(lr){5-6} \cmidrule(lr){7-8} \cmidrule(lr){9-10} \cmidrule(lr){11-12} \cmidrule(lr){13-14} \cmidrule(lr){15-16}
7 & 0.099$_{\texttt{10}}$ & 1 & 0.260$_{\texttt{10}}$ & 189 & 0.053$_{\texttt{0}}$ & 25 & 0.066$_{\texttt{10}}$ & 24 & 0.091$_{\texttt{10}}$ & 2 & 0.389$_{\texttt{10}}$ & 5 & 0.129$_{\texttt{10}}$ & 27 & 0.046$_{\texttt{10}}$ \\
6 & 0.098$_{\texttt{10}}$ & 2 & 0.196$_{\texttt{10}}$ & 112 & 0.043$_{\texttt{10}}$ & 651 & 0.054$_{\texttt{9}}$ & 14 & 0.065$_{\texttt{4}}$ & 3 & 0.312$_{\texttt{10}}$ & 4 & 0.084$_{\texttt{10}}$ & 34 & 0.041$_{\texttt{10}}$ \\
8 & 0.093$_{\texttt{10}}$ & 5 & 0.122$_{\texttt{10}}$ &  &  & 119 & 0.041$_{\texttt{7}}$ & 20 & 0.059$_{\texttt{3}}$ & 4 & 0.274$_{\texttt{10}}$ & 3 & 0.082$_{\texttt{10}}$ &  &  \\
10 & 0.084$_{\texttt{10}}$ & 13 & 0.080$_{\texttt{10}}$ &  &  &  &  & 22 & 0.053$_{\texttt{4}}$ & 5 & 0.254$_{\texttt{10}}$ & 19 & 0.058$_{\texttt{10}}$ &  &  \\
26 & 0.064$_{\texttt{10}}$ & 11 & 0.071$_{\texttt{10}}$ &  &  &  &  & 63 & 0.052$_{\texttt{4}}$ & 6 & 0.197$_{\texttt{10}}$ & 2 & 0.052$_{\texttt{10}}$ &  &  \\
12 & 0.061$_{\texttt{7}}$ & 21 & 0.067$_{\texttt{10}}$ &  &  &  &  & 345 & 0.051$_{\texttt{3}}$ & 8 & 0.114$_{\texttt{10}}$ & 34 & 0.036$_{\texttt{10}}$ &  &  \\
27 & 0.054$_{\texttt{10}}$ &  &  &  &  &  &  & 791 & 0.051$_{\texttt{3}}$ & 7 & 0.098$_{\texttt{10}}$ &  &  &  &  \\
56 & 0.052$_{\texttt{8}}$ &  &  &  &  &  &  & 1436 & 0.051$_{\texttt{3}}$ & 10 & 0.090$_{\texttt{10}}$ &  &  &  &  \\
36 & 0.051$_{\texttt{7}}$ &  &  &  &  &  &  & 173 & 0.051$_{\texttt{3}}$ & 14 & 0.088$_{\texttt{10}}$ &  &  &  &  \\
443 & 0.049$_{\texttt{0}}$ &  &  &  &  &  &  & 817 & 0.049$_{\texttt{2}}$ & 9 & 0.060$_{\texttt{10}}$ &  &  &  &  \\
33 & 0.047$_{\texttt{2}}$ &  &  &  &  &  &  & 1228 & 0.048$_{\texttt{3}}$ &  &  &  &  &  &  \\
1223 & 0.047$_{\texttt{0}}$ &  &  &  &  &  &  & 898 & 0.040$_{\texttt{1}}$ &  &  &  &  &  &  \\
1655 & 0.045$_{\texttt{2}}$ &  &  &  &  &  &  & 730 & 0.039$_{\texttt{2}}$ &  &  &  &  &  &  \\
1716 & 0.045$_{\texttt{0}}$ &  &  &  &  &  &  &  &  &  &  &  &  &  &  \\
505 & 0.038$_{\texttt{1}}$ &  &  &  &  &  &  &  &  &  &  &  &  &  &  \\
\bottomrule
\end{tabular}

    \caption{Mean amplitudes ($\amp_\component$) of the twenty-five most dominant sinusoids persistent across all folds after $L_1$ regularization. Subscripts denote the number of cross-validation folds (out of ten folds) where the sinusoids are significant. 
    \looseness-1}
    \label{tab:amplitudes-app}
\end{table*}

\clearpage

\end{document}